\documentclass{article}
\usepackage{booktabs}
\usepackage{multirow}
\usepackage{graphicx}
\usepackage{microtype}
\usepackage{graphicx}
\usepackage{subcaption}
\usepackage{booktabs} 
\usepackage[framemethod=default]{mdframed}
\usepackage{hyperref}
\usepackage{enumitem}

\usepackage[table]{xcolor}  
\usepackage{xspace} 

\usepackage[accepted]{icml2026}
\usepackage{makecell}
\usepackage{amsmath}
\usepackage{amssymb}
\usepackage{mathtools}
\usepackage{amsthm}
\newcommand{\Svl}{S_{\text{vl}}}
\newcommand{\Utxt}{U_{\text{txt}}}
\newcommand{\Uvis}{U_{\text{vis}}}
\newcommand{\Rvl}{R_{\text{vl}}}

\newcommand{\SR}{\text{SR}}
\newcommand{\SC}{\text{SC}}

\newcommand{\eg}{\emph{e.g.}\xspace}

\usepackage[capitalize,noabbrev]{cleveref}

\theoremstyle{plain}

\theoremstyle{definition}

\theoremstyle{remark}

\usepackage[textsize=tiny]{todonotes}

\icmltitlerunning{Towards Understanding Modality Interaction in Multimodal Language Models via Partial Information Decomposition}

\begin{document}

\twocolumn[
  \icmltitle{Towards Understanding Modality Interaction in Multimodal Language Models via Partial Information Decomposition}

\begin{icmlauthorlist}
  \icmlauthor{Wanlong Fang}{aix,ccds}
  \icmlauthor{Tianle Zhang}{ccds}
  \icmlauthor{Wen Tao}{ccds}
  \icmlauthor{Alvin Chan}{ccds,lkc,caim}
\end{icmlauthorlist}

\icmlaffiliation{aix}{Artificial Intelligence-X (AI-X), Interdisciplinary Graduate Programme, Nanyang Technological University, Singapore}
\icmlaffiliation{ccds}{College of Computing and Data Science, Nanyang Technological University, Singapore}
\icmlaffiliation{lkc}{Lee Kong Chian School of Medicine, Nanyang Technological University, Singapore}
\icmlaffiliation{caim}{Centre of AI in Medicine (C-AIM), Nanyang Technological University, Singapore}

\icmlcorrespondingauthor{Alvin Chan}{guoweialvin.chan@ntu.edu.sg}
  \icmlkeywords{Machine Learning, ICML}

  \vskip 0.3in
]



\printAffiliationsAndNotice{}  
\begin{abstract}
Understanding how multimodal large language models use different
modalities is important for reliable reasoning. We employ Partial
Information Decomposition (PID) as a decision-level lens and introduce
\emph{Sensory PID}, a conditional formulation that conditions on
language and separates unique, redundant, and synergistic contributions
from video and audio. Applied to omni-modal models, Sensory PID reveals
a sensory synergy bottleneck: even on audio--visual fusion tasks,
decisions remain dominated by modality-unique information, with stronger
reliance on vision. Modality-shuffling interventions support
this asymmetry, while layer-wise analysis reveals a visual-first
computation pattern and instruction perturbations show that late-stage
sensory fusion is conditioned by language. Beyond diagnosis, PID-guided
sample reweighting provides initial evidence that local diagnostic
signals can improve multimodal reasoning and grounding performance.
As reference validation, our vision--language analysis broadly
corroborates previously reported decision-level PID patterns across
tasks, models, interventions, and layers.
\end{abstract}

\section{Introduction}

\begin{figure*}
    \centering
    \includegraphics[width=0.9\linewidth]{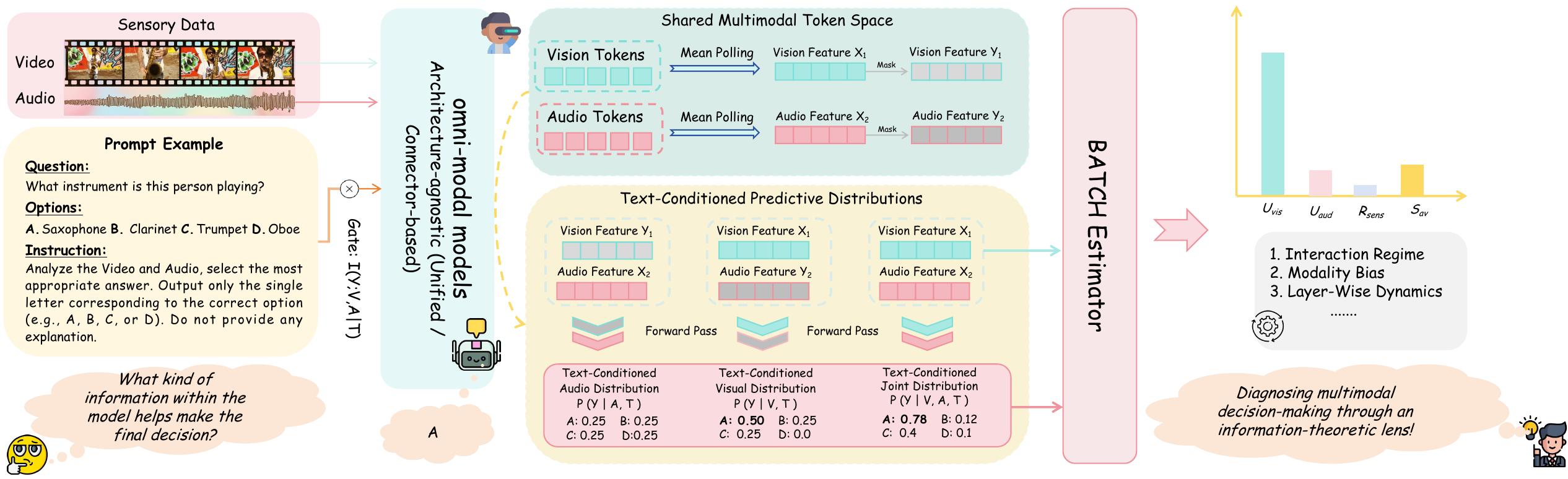}
    \caption{\textbf{The Sensory PID Framework.}
(Left) Omni-modal models are analyzed by treating the text prompt as a conditioning signal for video--audio interaction.
(Right) Applying the BATCH estimator in the shared token space decomposes decision-level information into Unique ($U_{\mathrm{vis}}, U_{\mathrm{aud}}$), Redundant ($R_{\mathrm{sens}}$), and Synergistic ($S_{\mathrm{av}}$) components.}
    \label{sensorypid}
    
\end{figure*}

Multimodal large language models (MLLMs) are moving from perception systems toward \emph{decision-making agents} in scientific analysis, medical support, and embodied interaction~\cite{yin2024survey,jin2025comprehensive,wang2025multimodal}. By integrating vision, audio, and language within unified generative backbones, these models can ground abstract instructions in rich sensory inputs and produce context-sensitive responses~\cite{liang2024comprehensive,deng2025survey,zhou2022learning}. Yet current evaluation paradigms still primarily assess \emph{what} models predict, while providing limited insight into \emph{how} different modalities contribute to those predictions~\cite{pan2024chain,deitke2025molmo}.

A growing body of work has sought to go beyond outcome accuracy through representation analysis, cross-modal alignment, attention-based inspection, and modality ablation~\cite{DBLP:journals/corr/abs-2502-14888,lou2025sae,yoon2025visual,jiang2025exploring,li2025vista}. These approaches provide complementary views of how modalities are encoded and how predictions respond to perturbations. Partial Information Decomposition (PID) offers another perspective by separating multimodal information into unique, redundant, and synergistic components. Our prior work used PID to characterize multimodal information structure and inform representation alignment~\cite{fang2026align}. More recently, \citet{xiu2026a} take a first step to extend PID to decision-level analysis of large vision--language models, revealing systematic information-use patterns across tasks, model families, interventions, and network depth.

These developments motivate a broader question: how do MLLMs use
different modalities when those modalities play distinct roles in
decision-making? This question becomes especially important in
omni-modal systems, where language specifies what the model should
resolve while video and audio provide sensory evidence. We therefore
introduce \emph{Sensory PID} (Figure~\ref{sensorypid}), a conditional
decision-level decomposition of the sensory information gain
$I(Y;V,A\mid T)$. By conditioning on language, Sensory PID separates
video-unique, audio-unique, redundant, and synergistic information
while distinguishing task specification from sensory evidence.

Applied to omni-modal models, Sensory PID reveals a \emph{sensory
synergy bottleneck}: even on questions designed to require joint
audio--visual reasoning, decision-relevant information remains
dominated by modality-unique contributions. Independent video and
audio shuffling supports a visually dominant decision profile.
Layer-wise analysis further reveals a \emph{visual-first} pattern in
which visual-unique information becomes decision-relevant before
auditory information and audiovisual synergy, while controlled
instruction perturbations show that late-stage sensory fusion varies
with linguistic task demand.

Beyond diagnosis, we ask whether PID-derived signals can guide model
adaptation. In a separate vision--language proof of concept, motivated
by our prior use of PID to inform multimodal intervention
design~\cite{fang2026align}, we use BATCH-induced local diagnostics to
reweight samples during lightweight LoRA fine-tuning. The policy
upweights samples with high fusion potential but limited current
synergy and downweights language-shortcut samples. The results provide
initial evidence that local information-decomposition signals can
improve reasoning and grounding over uniform fine-tuning and matched
reweighting baselines, while the contrast across model profiles
suggests that the benefit is profile-dependent.

For reference validation, we also apply the analogous two-source
decision-level PID analysis to 20 vision--language models across six
benchmarks. These experiments broadly corroborate previously reported
task-, model-, intervention-, and layer-dependent patterns
\citep{xiu2026a}, while extending the reference evaluation to MMStar, MMMU, and Qwen3-VL.

We summarize our main contributions as follows:

\paragraph{Conditional Sensory PID for omni-modal models.}
We introduce a conditional decomposition of $I(Y;V,A\mid T)$ that
treats language as task context and separates the decision-relevant
information contributed individually and jointly by video and audio.

\paragraph{Omni-modal sensory interaction findings.}
Across the evaluated omni-modal models, Sensory PID reveals a sensory
synergy bottleneck, intervention-supported visual dominance,
visual-first information dynamics, and instruction-conditioned sensory
fusion.

\paragraph{PID-guided sample reweighting.}
In a separate VLM proof of concept, we formulate a PID-guided
sample-reweighting policy that converts BATCH-induced local
information-decomposition diagnostics into sample rankings and
training weights. The resulting experiments provide initial evidence
of reasoning and grounding gains in joint-modality-reliant models
while exposing a profile-dependent boundary case.
\label{sec:method}

\section{Methodology}
\label{sec:method}

Our methodology has two complementary components. First, we introduce
\emph{Sensory PID}, a conditional decision-level decomposition for
characterizing how video and audio contribute to an omni-modal model's
decision under a textual task specification. Second, we formulate a
PID-guided sample-reweighting policy that converts local
information-decomposition diagnostics into sample-ranking and training
signals for model adaptation. The former provides an analytical
framework for language-conditioned sensory interaction, while the
latter translates information decomposition from diagnosis into a
targeted training intervention.

\subsection{Conditional Sensory PID for Omni-Modal Models}
\label{sec:sensory_pid}

Omni-modal models process a video stream $V$, an audio stream $A$,
and a textual instruction $T$ to produce a response $Y$.
Our goal is to characterize how the two sensory streams contribute
individually and jointly to the model's decision \emph{under a fixed
linguistic task specification}. A full three-source PID is possible
in principle, but introduces a combinatorial number of
partial-information terms and treats language symmetrically with the
sensory inputs. In instruction-following settings, however, these
modalities play different roles: we treat $T$ as the conditioning
context specifying the task, while $V$ and $A$ provide the sensory
evidence relevant to that task.

We therefore introduce \emph{Sensory PID}, which decomposes the
conditional sensory information gain
\begin{equation}
I(Y;V,A\mid T)
=
U_{\mathrm{vis}}
+
U_{\mathrm{aud}}
+
R_{\mathrm{sens}}
+
S_{\mathrm{av}},
\label{eq:sensory_pid}
\end{equation}
where $U_{\mathrm{vis}}$ and $U_{\mathrm{aud}}$ denote information
uniquely attributable to video and audio, respectively;
$R_{\mathrm{sens}}$ denotes sensory information redundantly available
from either stream; and $S_{\mathrm{av}}$ denotes information available
only from their joint observation, conditioned on the instruction.

The conditioning structure is central to the formulation. Rather than
treating text as a third PID source symmetric with video and audio,
Sensory PID keeps the instruction fixed and asks how the available
sensory evidence contributes to the resulting decision. This separates
task specification from sensory contribution and changes the object of
decomposition from generic multi-source interaction to
language-conditioned video--audio interaction.

Operationally, the analysis is defined through three model-induced,
text-conditioned predictive distributions:
\begin{equation}
\begin{gathered}
    p_\theta(Y\mid V,A,T),
    \qquad
    p_\theta(Y\mid V,T),
    \\
    p_\theta(Y\mid A,T).
\end{gathered}
\label{eq:sensory_conditionals}
\end{equation}
corresponding respectively to the full sensory prediction and the two
partial-sensory conditions. The same instruction $T$ is retained in
all three cases, so their differences reflect the sensory evidence
available to the model. Section~\ref{sec:sensory_estimation} describes
how these distributions and the resulting Sensory PID components are
estimated in practice.

\subsection{Estimating Sensory PID in Omni-Modal Models}
\label{sec:sensory_estimation}

Estimating Sensory PID in modern omni-modal models requires addressing
three practical issues: sensory inputs are represented by
variable-length token sequences, jointly trained models do not always
provide native partial-sensory inference paths under a fixed
instruction, and restricted candidate distributions can become
unstable when one sensory stream is removed. We therefore construct a
training-free estimation pipeline around the BATCH estimator
(Liang et al., 2023), as illustrated in Figure~\ref{sensorypid}.

\paragraph{Sensory source representations.}
Omni-modal architectures map video, audio, and language into a shared
multimodal representation space. Because BATCH operates on
fixed-dimensional continuous source variables, we summarize the video
and audio token sequences using mean pooling after their projection
into this shared space. This yields sensory source representations
$X_V,X_A\in\mathbb{R}^d$, while the textual instruction remains present
as conditioning context throughout inference. Mean pooling provides a
parameter-free, sample-aligned summary of each sensory stream;
sensitivity to alternative pooling strategies is examined in
Appendix~G.

\paragraph{Text-conditioned sensory predictions.}
For each sample $(V,A,T)$, the full omni-modal forward pass produces
$p_\theta(Y\mid V,A,T)$. To estimate the two partial-sensory
conditions required by Eq.~\ref{eq:sensory_pid}, we additionally
construct
$p_\theta(Y\mid V,T)$ and $p_\theta(Y\mid A,T)$ while keeping the same
instruction $T$. Across the profiling set
$\mathcal{D}_{\mathrm{prof}}$, these predictions define the
model-induced decision distributions used for Sensory PID estimation.

\paragraph{Calibrated sensory masking.}
Because current omni-modal models do not always expose native
video--text or audio--text execution paths, we approximate these
partial-sensory conditions through representation corruption. To
estimate $p_\theta(Y\mid V,T)$, we replace the audio embeddings while
retaining video and text; conversely, to estimate
$p_\theta(Y\mid A,T)$, we replace the video embeddings while retaining
audio and text. For a masked sensory modality $m\in\{V,A\}$,
\begin{equation}
\widetilde{E}_{m}[\ell]
\sim
\mathcal{N}\!\left(
\mu_m,\operatorname{diag}(\sigma_m^2)
\right),
\qquad
\ell=1,\ldots,L_m ,
\label{eq:sensory_masking}
\end{equation}
where $\mu_m$ and $\sigma_m$ are per-dimension statistics estimated
over $\mathcal{D}_{\mathrm{prof}}$. Matching the empirical feature
statistics controls the perturbation scale while removing
instance-specific information from the masked sensory stream.

\paragraph{Sensory PID estimation.}
We apply BATCH~\cite{liang2023quantifying} to the video and audio source representations together
with the text-conditioned decision distributions. BATCH learns a
Sinkhorn-normalized coupling satisfying the required source--target
marginal constraints, from which
$U_{\mathrm{vis}}$, $U_{\mathrm{aud}}$,
$R_{\mathrm{sens}}$, and $S_{\mathrm{av}}$ are recovered.
The instruction therefore specifies the decision context throughout
the estimation, while the decomposition characterizes how video and
audio contribute information within that context.

\paragraph{Output stabilization.}
For multiple-choice evaluation, each predictive distribution is
restricted to the candidate set $\mathcal{C}$ using selective softmax.
Prior decision-level LVLM PID analysis uses a hard confidence gate:
when the probability mass assigned to valid candidates falls below a
threshold $\tau$, the corresponding candidate distribution is replaced
by a uniform distribution before soft aggregation
~\citep{xiu2026a}.

This stabilization rule is not directly aligned with our conditional
sensory setting. Sensory PID jointly relies on the full-sensory
prediction $p_\theta(Y\mid V,A,T)$ and the two partial-sensory
predictions $p_\theta(Y\mid V,T)$ and
$p_\theta(Y\mid A,T)$. Because differences among these predictive
conditions directly enter the estimation of sensory-unique and
audiovisual-synergistic information, independently replacing a
distribution once its confidence crosses a hard threshold can introduce
an abrupt branch-specific change that is unrelated to the underlying
sensory evidence.

We therefore use a continuous confidence-dependent stabilization.
For each sample $i$ and predictive condition $c$, let
\begin{equation}
m_i^{(c)}
=
\sum_{y\in\mathcal{C}}
S_{\mathrm{orig},i}^{(c)}(y)
\end{equation}
denote the pre-renormalization probability mass assigned to valid
candidates. We define
\begin{equation}
\rho_i^{(c)}
=
\min\left\{
1,\,
\frac{m_i^{(c)}}{\tau}
\right\},
\label{eq:sensory_reliability}
\end{equation}
and stabilize the corresponding predictive distribution as
\begin{equation}
\widetilde{p}_i^{(c)}(y)
=
\rho_i^{(c)}p_i^{(c)}(y)
+
\left(1-\rho_i^{(c)}\right)
\frac{1}{|\mathcal{C}|}.
\label{eq:sensory_confidence_shrinkage}
\end{equation}
Predictions above the confidence threshold are therefore unchanged,
whereas lower-confidence predictions are progressively shrunk toward
the uniform distribution rather than being replaced discontinuously.

For the target marginal, we retain the soft-aggregation principle used
in prior decision-level LVLM PID analysis~\citep{xiu2026a}:
\begin{equation}
\bar{p}(y)
=
\frac{1}{|\mathcal{D}_{\mathrm{prof}}|}
\sum_{j\in\mathcal{D}_{\mathrm{prof}}}
\widetilde{p}_j^{(VAT)}(y),
\label{eq:sensory_soft_aggregation}
\end{equation}
and keep this marginal fixed during BATCH optimization
~\citep{liang2023quantifying}.
Thus, our implementation shares the soft-aggregation principle with
prior LVLM PID analysis but differs in how low-confidence predictive
conditions are stabilized before aggregation.
~\citep{liang2023quantifying}.
\paragraph{Estimator provenance and vision--language reference.}
BATCH, mean pooling, embedding-space corruption, confidence
regularization, and soft aggregation are implementation choices for
estimating PID quantities rather than methodological contributions of
this work. Closely related combinations of these choices have also
been reported in prior decision-level LVLM
analysis~\citep{xiu2026a}. Our contribution in the omni-modal setting
lies in formulating and applying conditional Sensory PID for
language-conditioned video--audio interaction, rather than in these
individual estimator components.

For the vision--language reference analysis and the reweighting method
introduced in Section~\ref{sec:pid_reweighting}, we use the analogous
two-source vision--language formulation; its definition and
implementation details are provided in
Appendix~\ref{app:vl_pid}. The underlying two-source VLM PID
formulation and estimator are not treated as methodological
contributions of this work.

\subsection{PID-Guided Reweighting from Local Diagnostics}
\label{sec:pid_reweighting}
\label{sec:guide_method}

\paragraph{From diagnosis to intervention.}
Beyond using information decomposition as an analytical tool, we ask
whether its local diagnostic signals can guide model adaptation.
Motivated by our prior use of PID to inform multimodal intervention
design~\cite{fang2026align}, we formulate a PID-guided
sample-reweighting policy for vision--language models. The policy is
designed to distinguish samples with unrealized cross-modal fusion
potential from those dominated by language-side shortcuts. Importantly,
the contribution here is not a new PID estimator, but the conversion
of BATCH-induced local diagnostic signals into a structured
sample-ranking and reweighting policy.

\paragraph{Local information-decomposition diagnostics.}
Standard sample-selection criteria conflate different sources of
difficulty. Accuracy-based mining identifies hard examples but cannot
distinguish weak multimodal fusion from missing knowledge, while
modality-ablation scores measure dependence on a modality without
separating unique from synergistic information.

For each training sample $i$, BATCH induces local additive
contributions $s_i$, $u_{\mathrm{txt},i}$,
$u_{\mathrm{vis},i}$, and $r_i$ whose averages recover the aggregate
PID terms. These quantities are used as BATCH-induced diagnostic
contributions rather than axiomatic pointwise PID atoms; their full
derivation and stabilization procedure are provided in
Appendix~\ref{app:local_scores}.

Because the local contributions need not be non-negative, we apply
non-negative clipping, $[z]_+=\max(z,0)$, and define the per-sample
information mass
\begin{equation}
\begin{aligned}
    I_i^{+}
    &=
    [s_i]_+
    +
    [u_{\mathrm{vis},i}]_+
    +
    [u_{\mathrm{txt},i}]_+
    +
    [r_i]_+ .
\end{aligned}
\label{eq:information_mass}
\end{equation}
We then define the \emph{synergy ratio} and
\emph{shortcut score}
\begin{equation}
    \SR_i
    =
    \frac{[s_i]_+}{I_i^{+}+\epsilon},
    \qquad
    \SC_i
    =
    \frac{[u_{\mathrm{txt},i}]_+}{I_i^{+}+\epsilon},
    \label{eq:sr_sc}
\end{equation}
where $\epsilon=10^{-8}$. To reduce dependence on individual
mini-batch composition, the local contributions are averaged over
$K=50$ random batch contexts before constructing these scores.

\paragraph{Fusion-potential and synergy-gap scoring.}
The frozen VLM additionally provides three predictive distributions:
$p_v^{(i)}=P(Y\mid V_i)$,
$p_t^{(i)}=P(Y\mid T_i)$, and
$p_{vt}^{(i)}=P(Y\mid V_i,T_i)$.
We define the \emph{fusion potential}
\begin{equation}
    \mathrm{FP}_i
    =
    \left[
    \min
    \left\{
        H(p_v^{(i)}),
        H(p_t^{(i)})
    \right\}
    -
    H(p_{vt}^{(i)})
    \right]_+ ,
    \label{eq:fp}
\end{equation}
which measures how much the joint prediction sharpens relative to the
better of the two partial-input predictions.

Fusion potential alone does not indicate whether the model already
uses the two modalities synergistically. We therefore combine it with
the local PID profile and define
\begin{equation}
    \mathrm{GapScore}_i
    =
    (1-\SR_i)(1-\SC_i)\,
    \mathrm{FP}_i ,
    \label{eq:gapscore}
\end{equation}
which prioritizes samples with high apparent fusion potential, limited
current synergy, and limited text-unique shortcut reliance.

\paragraph{PID-guided sample reweighting.}
The resulting policy has two complementary priorities. Samples with
the largest $\SC_i$ are treated as \emph{language-shortcut} examples
and downweighted, while samples with the largest $\mathrm{GapScore}_i$
among the remaining examples are treated as \emph{synergy-gap}
examples and upweighted:
\begin{equation}
\begin{aligned}
    \mathcal{S}_{\mathrm{short}}
    &=
    \operatorname{TopK}_{K_s}(\SC_i),
    \\
    \mathcal{S}_{\mathrm{gap}}
    &=
    \operatorname{TopK}_{K_g}
    \bigl(
    \mathrm{GapScore}_i
    \mid i\notin\mathcal{S}_{\mathrm{short}}
    \bigr).
\end{aligned}
\label{eq:sample_selection}
\end{equation}
We assign
$w_i=3.0$ to synergy-gap samples,
$w_i=0.5$ to language-shortcut samples, and
$w_i=1.0$ otherwise.
The resulting weights are used during lightweight LoRA fine-tuning.
Algorithmic details and the effective training distribution are
reported in Appendix~\ref{app:reweighting}.
\section{Experimental Setup}
\label{sec:setup}

Our evaluation is organized around two complementary roles. The
\emph{primary} experiments study language-conditioned sensory
interaction in omni-modal models using Sensory PID, including
modality-specific interventions, layer-wise analysis, and controlled
instruction perturbations. Separately, we retain a vision--language
evaluation as a reference check for the decision-level PID pipeline.
As this reference analysis primarily corroborates previously
reported LVLM patterns~\citep{xiu2026a}, we keep its presentation
concise in the main text and defer the detailed results to appendix.

\paragraph{Omni-modal models.}
We evaluate three model configurations from two omni-modal families:
Qwen2.5-Omni 3B and 7B~\citep{xu2025qwen2}, and VITA-1.5
7B~\citep{fu2025vita}. These models support joint video, audio, and
text inputs and provide coverage across distinct model families and
parameter scales.

\paragraph{Omni-modal benchmark and sensory subsets.}
Our primary omni-modal evaluation uses MUSIC-AVQA~\citep{li2022learning},
an audio--visual question-answering benchmark based on music videos.
To distinguish sensory requirements, we construct three
balanced subsets from the dataset metadata:
\emph{Audio-Focused}, containing questions whose answers depend
primarily on auditory evidence;
\emph{Visual-Focused}, emphasizing visual objects or actions; and
\emph{AV-Fusion}, containing questions designed to require joint
audio--visual reasoning.
The AV-Fusion subset serves as our primary testbed for sensory synergy,
while the Audio-Focused and Visual-Focused subsets provide reference
conditions for modality-specific information use. Detailed dataset
statistics and construction criteria are provided in
Appendix~\ref{app:datasets}.

\paragraph{Sensory evaluation and intervention protocol.}
All models are evaluated zero-shot using standardized prompts and
deterministic greedy decoding. For Sensory PID estimation, the same
textual instruction is retained across the full and partial-sensory
conditions described in \S\ref{sec:sensory_pid}, so that the decomposition
compares sensory evidence under a fixed linguistic task context.

We complement the information decomposition with three analyses that
probe different aspects of sensory interaction. First, we independently
shuffle video or audio streams across samples, breaking the
correspondence of one sensory modality while preserving its
sample-level marginal distribution. This provides a behavioral measure
of the model's sensitivity to correctly aligned visual and auditory
evidence. Second, we trace Sensory PID components across transformer
depth to examine when video-, audio-, and jointly derived information
become decision-relevant. Third, we hold the video and audio inputs
fixed while modifying the textual instruction to reduce its demand for
joint audio--visual reasoning, allowing us to examine whether sensory
fusion changes with linguistic task specification.

\paragraph{Vision--language reference evaluation.}
As a separate reference validation, we evaluate 20 vision--language
models from 7 model families across six benchmarks: MMBench,
MMStar, MMMU, PMC-VQA, POPE, and Reefknot. We additionally obtain a
text-only condition by removing the image input and measure
$\Delta_{\text{vision}}$ as the corresponding accuracy drop.
These experiments serve as a reference check and broadly recover the
task-, model-, intervention-, and layer-dependent PID patterns reported
in prior LVLM analysis~\citep{xiu2026a}, rather than constituting the
primary empirical contribution of this work. Benchmark details and
full per-model results are provided in
Appendices~\ref{app:datasets} and~\ref{app:full_vl}.
The separate training setup for the PID-guided reweighting experiments
is described in \S\ref{sec:guide}.

\section{Sensory Interaction in Omni-Modal Models}
\label{sec:results}

\paragraph{Vision--language reference validation.}
As a reference check, we also apply the analogous decision-level PID
analysis to 20 vision--language models across six benchmarks.
The resulting task-, model-, intervention-, and layer-dependent
patterns are broadly consistent with prior LVLM PID
analysis~\cite{xiu2026a}: benchmarks separate into relatively
synergy-heavy and text-unique-heavy regimes, family-level modality-use
tendencies are reflected in vision-removal sensitivity, and the
previously reported late-stage shift toward cross-modal synergy also
appears in our layer-wise analyses on MMStar and MMMU. Because these
analyses primarily serve as reference validation, we defer the complete
results, correlations, and layer-wise trajectories to
Appendix~\ref{app:full_vl} and focus below on sensory interaction in
omni-modal models.

\begin{table*}[t]
\centering
\caption{\textbf{Sensory PID spectrum and performance across Audio-Focused, Visual-Focused, and AV-Fusion subsets.}
$U_{\text{vis}}$, $U_{\text{aud}}$, $S_{\text{av}}$, and $R_{\text{sens}}$ denote unimodal visual contribution, unimodal audio contribution, audiovisual synergy, and sensory redundancy, respectively, computed via conditional PID on $I(Y; V, A \mid T)$. The dominant information component for each model–subset pair is highlighted in \textbf{bold}. $Acc$ and $Acc1$ report full audio-visual accuracy and visual-only accuracy under audio removing. }
\label{table3}
\renewcommand{\arraystretch}{1.2}
\setlength{\tabcolsep}{1.2pt} 
\scriptsize
\resizebox{\textwidth}{!}{%
\begin{tabular}{@{}l cccccc c cccccc c cccccc c cccccc@{}}
\toprule
\multirow{2}{*}{\textbf{Model}} & \multicolumn{6}{c}{\textbf{Audio-Focused}} && \multicolumn{6}{c}{\textbf{Visual-Focused}} && \multicolumn{6}{c}{\textbf{AV-Fusion}} && \multicolumn{6}{c}{\textbf{Full Selection}} \\
\cmidrule{2-7} \cmidrule{9-14} \cmidrule{16-21} \cmidrule{23-28}
 & $U_{\text{vis}}$ & $U_{\text{aud}}$ & $S_{\text{av}}$ & $R_{\text{sens}}$ & $Acc$ & $Acc1$ && $U_{\text{vis}}$ & $U_{\text{aud}}$ & $S_{\text{av}}$ & $R_{\text{sens}}$ & $Acc$ & $Acc1$ && $U_{\text{vis}}$ & $U_{\text{aud}}$ & $S_{\text{av}}$ & $R_{\text{sens}}$ & $Acc$ & $Acc1$ && $U_{\text{vis}}$ & $U_{\text{aud}}$ & $S_{\text{av}}$ & $R_{\text{sens}}$ & $Acc$ & $Acc1$ \\ \midrule
VITA-1.5 7B & 0.54 & \textbf{0.59} & 0.20 & 0.005 & 51.6 & 25.4 && \textbf{1.34} & 0.25 & 0.15 & 0.005 & 77.5 & 75.8 && \textbf{1.35} & 0.60 & 0.22 & 0.004 & 57.8 & 56.2 && \textbf{1.09} & 0.55 & 0.22 & 0.003 & 61.4 & 54.2 \\
Qwen2.5-Omni 3B & 0.48& \textbf{0.64} & 0.15 & 0.004 & 48.0 & 22.0 && \textbf{1.26} & 0.22 & 0.12 & 0.004 & 70.0 & 68.0 && \textbf{1.25} & 0.55 & 0.21 & 0.003 & 52.0 & 50.0 && \textbf{1.06} & 0.50 & 0.18 & 0.001 & 55.0 & 48.0 \\
Qwen2.5-Omni 7B & \textbf{0.71} & 0.69 & 0.26 & 0.001 & 55.2 & 42.1 && \textbf{1.36} & 0.28 & 0.17 & 0.002 & 79.0 & 77.0 && \textbf{1.42} & 0.65 & 0.32 & 0.004 & 62.0 & 58.0 && \textbf{1.12} & 0.58 & 0.24 & 0.004 & 64.0 & 56.0 \\

\bottomrule
\end{tabular}%
 }
 
\end{table*}


\subsection{Sensory Synergy Bottleneck and Visual Dominance}
\label{sec:omni_visual}

We first examine how video and audio contribute to decisions on the
AV-Fusion subset of MUSIC-AVQA, where questions are designed to
emphasize joint audio--visual reasoning. If the two sensory streams
are strongly integrated at the decision level, these questions should
exhibit substantial sensory synergy. Instead, Sensory PID reveals a
pronounced imbalance between modality-unique and jointly available
sensory information.

\subsubsection{The Sensory Synergy Bottleneck}
\label{sec:sensory_bottleneck}

Table~\ref{table3} shows that sensory synergy $S_{\text{av}}$ remains
a relatively small component across the evaluated omni-modal models,
even on the AV-Fusion subset. Rather than deriving most
decision-relevant sensory information from the joint observation of
video and audio, the models remain dominated by modality-unique
information. We refer to this pattern as a \emph{sensory synergy
bottleneck}: supporting multiple sensory streams architecturally does
not necessarily yield strong decision-level audiovisual integration.

The decomposition also reveals a consistent asymmetry between the two
sensory modalities. Unique visual information $U_{\text{vis}}$
generally exceeds unique auditory information $U_{\text{aud}}$,
indicating stronger reliance on visual evidence. This visual skew
motivates the intervention analysis below, where we test whether the
information imbalance is accompanied by asymmetric sensitivity to
disrupting the two sensory streams.

\begin{mdframed}[
backgroundcolor=gray!5,
linewidth=0.5pt,
innertopmargin=4pt,
innerbottommargin=4pt
]
\textbf{Finding 1.}
Across the evaluated omni-modal models, even questions designed to
require joint audio--visual reasoning derive most decision-relevant
sensory information from modality-unique signals, while sensory
synergy remains limited. This reveals a \emph{sensory synergy
bottleneck} in the evaluated systems.
\end{mdframed}


\subsubsection{Functional Validation via Modality Shuffling}

To examine whether this sensory imbalance is reflected in observable
model behavior, we independently shuffle video or audio streams across
samples. The intervention breaks the correspondence between the
question and one sensory stream while preserving its modality-specific
marginal distribution.

\begin{figure}[t]
\centering
\includegraphics[width=0.9\linewidth]{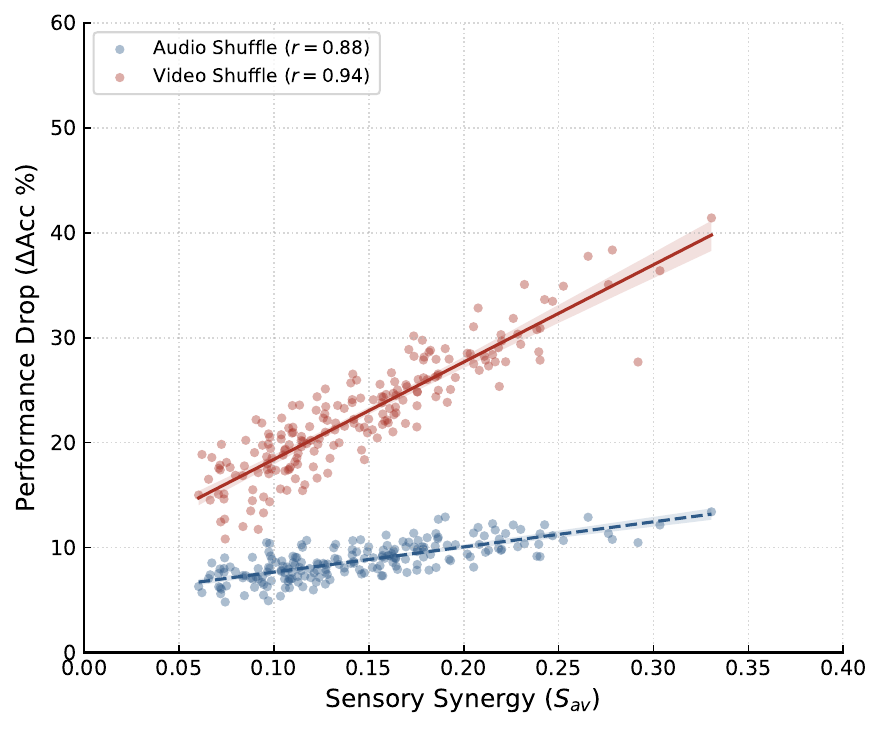}
\caption{\textbf{Functional validation via modality shuffling.}
Correlation between Sensory Synergy ($S_{\text{av}}$) and accuracy
drop ($\Delta \mathrm{Acc}$) on the AV-Fusion subset of MUSIC-AVQA
(Qwen2.5-Omni 7B) under audio or video shuffling. Each point represents
a question instance; Spearman $\rho$ is reported.}
\vspace{-15pt}
\label{figure2}
\end{figure}

Consistent with the Sensory PID spectrum in Table~\ref{table3},
shuffling the video stream induces substantially larger performance
drops than shuffling the audio stream, even on AV-Fusion questions.
This asymmetric sensitivity indicates that model predictions depend
more strongly on correctly aligned visual information than on
auditory information under these modality-level perturbations.

\begin{mdframed}[
backgroundcolor=gray!5,
linewidth=0.5pt,
innertopmargin=4pt,
innerbottommargin=4pt
]
\textbf{Finding 2.}
Modality-shuffling interventions support a visually dominant decision
profile: the evaluated omni-modal models suffer larger performance
drops under visual than auditory misalignment, while Sensory PID is
associated with sensitivity to cross-modal disruption.
\end{mdframed}


\subsection{Layer-Wise Sensory Dynamics}
\label{sec:layerwise}

The final-layer sensory imbalance raises a further question: when does
this asymmetry emerge during computation? We therefore trace Sensory
PID across network depth to examine the temporal organization of
video-, audio-, and jointly derived decision information.

\begin{figure*}[t]
    \centering
    \includegraphics[width=0.99\linewidth]{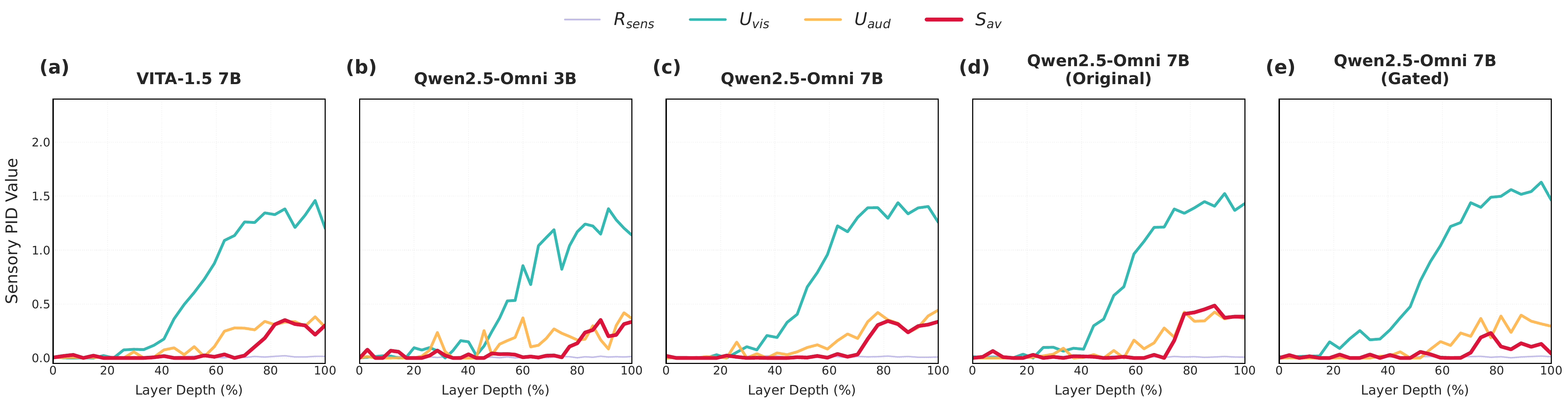}
    \caption{\textbf{Layer-wise Sensory PID and instruction-conditioned
    fusion in omni-modal models.}
    Sensory PID across depth for VITA-1.5 and Qwen2.5-Omni on the
    AV-Fusion subset of MUSIC-AVQA. Panels (a--c) plot
    $U_{\text{vis}}$, $U_{\text{aud}}$, $R_{\text{sens}}$, and
    $S_{\text{av}}$ versus normalized depth. Panels (d--e) illustrate
    the effect of replacing a fusion-demanding instruction with a
    less fusion-dependent alternative while holding the sensory inputs
    fixed.}
    \vspace{-11pt}
    \label{figure4}
\end{figure*}


\subsubsection{Visual-First Dynamics in Omni-Modal Models}
\label{sec:visual_first}

Figure~\ref{figure4}(a)--(c) reveals a pronounced asymmetry in the
temporal emergence of video-, audio-, and jointly derived information.
Although the overall progression moves from weak early information
toward stronger multimodal information in later layers, the two sensory
streams do not become decision-relevant at comparable rates.

\paragraph{Early visual dominance.}
$U_{\text{vis}}$ rises substantially earlier and more sharply than
$U_{\text{aud}}$, becoming the dominant sensory component through much
of the intermediate computation. Auditory-unique information remains
comparatively limited during this stage. This ordering indicates that
visual evidence becomes decision-relevant earlier, producing an
already visually skewed predictive state before substantial auditory
or cross-sensory information emerges.

\paragraph{Constrained late sensory integration.}
Sensory synergy $S_{\text{av}}$ increases primarily in later layers,
indicating that audiovisual integration is delayed relative to the
accumulation of visual-unique information. Although this late-stage
rise indicates increasing joint use of the two sensory streams, its
magnitude generally remains below $U_{\text{vis}}$. Together with the
final-layer decomposition in Table~\ref{table3}, these dynamics suggest
that the sensory synergy bottleneck is not simply a final-output
phenomenon: it is associated with a temporal imbalance in which visual
information becomes decision-relevant earlier than auditory and
synergistic information.

\paragraph{Architecture and scale modulate timing, but not the ordering.}
The precise trajectories vary across model families. VITA-1.5 exhibits
an earlier and steeper increase in $U_{\text{vis}}$, whereas
Qwen2.5-Omni shows a more gradual rise before converging to a similarly
visual-dominant profile in later layers. Model scale also affects the
smoothness and magnitude of these trajectories. Across the evaluated
architectures and scales, however, the qualitative ordering remains
consistent: visual-unique information emerges first, while
auditory-unique information and audiovisual synergy become prominent
later.

\begin{mdframed}[
backgroundcolor=gray!5,
linewidth=0.5pt,
innertopmargin=4pt,
innerbottommargin=4pt
]
\textbf{Finding 3.}
Across the evaluated omni-modal models, unique visual information
emerges earlier than both unique auditory information and audiovisual
synergy. This consistent temporal ordering reveals a
\emph{visual-first computation pattern}: intermediate decisions become
visually dominated before stronger cross-sensory integration develops
in later layers.
\end{mdframed}


\subsubsection{Instruction-Conditioned Sensory Fusion}
\label{sec:instruction_gating}

We finally examine whether the late-stage audiovisual synergy observed
above varies with the linguistic task specification. Holding the video
and audio streams fixed, we modify the instruction to reduce its demand
for joint audiovisual reasoning and re-estimate the layer-wise Sensory
Synergy $S_{\text{av}}$.

Figure~\ref{figure4}(d)--(e) provides an illustrative example.
Replacing a fusion-demanding query such as
``\textit{Is the violin playing the high pitch?}''
with a less fusion-dependent alternative such as
``\textit{Which instrument is being played?}''
substantially attenuates late-layer $S_{\text{av}}$, while leaving the
early- and mid-layer unimodal trajectories comparatively stable.
Thus, identical sensory inputs can exhibit different integration
profiles depending on the task specified by the language instruction.

This observation complements the visual-first pattern above: sensory
evidence begins to accumulate before substantial audiovisual synergy
emerges, while the extent of late-stage fusion varies with the
instructional demand for joint sensory reasoning. Language therefore
serves as a task-conditioning context for the sensory interaction
analyzed by Sensory PID.

\begin{mdframed}[
backgroundcolor=gray!5,
linewidth=0.5pt,
innertopmargin=4pt,
innerbottommargin=4pt
]
\textbf{Finding 4.}
Holding video and audio fixed, changing the instruction alters
late-stage audiovisual synergy while leaving earlier unimodal dynamics
comparatively stable. This provides evidence for
\emph{instruction-conditioned sensory fusion}: the extent to which
joint sensory evidence becomes decision-relevant varies with the
linguistic task demand.
\end{mdframed}

\section{Evaluating PID-Guided Reweighting}
\label{sec:guide}



We now evaluate the PID-guided sample-reweighting policy introduced in
Section~\ref{sec:pid_reweighting}. These experiments serve as a
separate vision--language proof of concept for testing whether local
information-decomposition diagnostics provide useful training guidance
beyond uniform fine-tuning, generic sample difficulty, or simple
modality sensitivity. We evaluate both downstream performance and the
post-tuning PID profile to examine where the proposed policy is
effective and where its benefits are limited.

\begin{table*}[t]
\centering
\caption{\textbf{PID-guided reweighting results on Qwen2.5-VL-7B.}
Values are mean$\pm$std over three seeds.
Post-$\Svl$ and Post-$\Utxt$ are estimated on a held-out MMStar
subset.}
\label{tab:main_rw}
\vspace{-0.4em}

\small
\setlength{\tabcolsep}{5pt}
\begin{tabular*}{\textwidth}{@{\extracolsep{\fill}}lccccccc@{}}
\toprule
& \multicolumn{2}{c}{\textbf{Reasoning}}
& \textbf{Grounding}
& \multicolumn{2}{c}{\textbf{Prior-Dominated}}
& \multicolumn{2}{c}{\textbf{PID Diagnostics}} \\
\cmidrule(lr){2-3}
\cmidrule(lr){4-4}
\cmidrule(lr){5-6}
\cmidrule(lr){7-8}

\textbf{Condition}
& \textbf{MMBench}
& \textbf{MMStar}
& \textbf{POPE}
& \textbf{MMMU}
& \textbf{PMC-VQA}
& \textbf{Post-$\Svl$}
& \textbf{Post-$\Utxt$} \\
\midrule

A: Base
& 88.3
& 60.7
& 86.4
& 54.1
& 53.1
& 1.15
& 0.60 \\

B: LoRA-Uniform
& 89.1{\tiny$\pm$.3}
& 62.0{\tiny$\pm$.4}
& 87.2{\tiny$\pm$.3}
& 54.0{\tiny$\pm$.3}
& 53.0{\tiny$\pm$.3}
& 1.20{\tiny$\pm$.02}
& 0.56{\tiny$\pm$.02} \\

\textbf{C: LoRA-PID}
& \textbf{90.2}{\tiny$\pm$.2}
& \textbf{64.3}{\tiny$\pm$.3}
& \textbf{88.5}{\tiny$\pm$.2}
& 53.5{\tiny$\pm$.4}
& 52.7{\tiny$\pm$.3}
& \textbf{1.36}{\tiny$\pm$.03}
& \textbf{0.46}{\tiny$\pm$.02} \\

D: LoRA-Random
& 88.8{\tiny$\pm$.5}
& 61.5{\tiny$\pm$.6}
& 86.9{\tiny$\pm$.4}
& 54.1{\tiny$\pm$.4}
& 53.2{\tiny$\pm$.4}
& 1.16{\tiny$\pm$.04}
& 0.58{\tiny$\pm$.03} \\

E: LoRA-Acc
& 89.4{\tiny$\pm$.3}
& 62.5{\tiny$\pm$.4}
& 87.5{\tiny$\pm$.3}
& \textbf{54.3}{\tiny$\pm$.3}
& \textbf{53.4}{\tiny$\pm$.3}
& 1.22{\tiny$\pm$.02}
& 0.54{\tiny$\pm$.02} \\

F: LoRA-Ablation
& 89.6{\tiny$\pm$.3}
& 63.0{\tiny$\pm$.4}
& 87.8{\tiny$\pm$.3}
& 53.8{\tiny$\pm$.3}
& 52.9{\tiny$\pm$.3}
& 1.24{\tiny$\pm$.02}
& 0.52{\tiny$\pm$.02} \\
\bottomrule
\end{tabular*}

\vspace{-8pt}
\end{table*}

\subsection{Experimental Setup and Main Results}
\label{sec:guide_results}

\paragraph{Experimental setup.}
We profile 3{,}000 MMBench training samples and compare LoRA-PID
against standard uniform LoRA and three matched-count reweighting
controls: random assignment, accuracy-based difficulty mining, and
modality-ablation-based selection. The primary experiment uses
Qwen2.5-VL-7B, with additional transfer experiments on
LLaVA-OneVision-7B and Gemma3-12B. Together, these controls test
whether the PID-guided policy provides information beyond non-uniform
weighting, generic sample difficulty, or simple sensitivity to visual
input.

All reweighting conditions use identical sample counts and LoRA
hyperparameters. Adapters are placed in the final 20\% of transformer
layers. No evaluation samples are used for PID profiling,
hyperparameter selection, or LoRA training. Results are reported over
three random seeds, and post-tuning PID diagnostics are estimated on a
held-out MMStar subset. Full experimental conditions, data controls,
and LoRA configurations are provided in
Appendix~\ref{app:reweighting}.

\paragraph{PID-guided reweighting outperforms matched alternatives.}
Table~\ref{tab:main_rw} shows that LoRA-PID improves all three
reasoning- and grounding-oriented benchmarks relative to uniform
fine-tuning: $+1.1$~pp on MMBench, $+2.3$~pp on MMStar, and
$+1.3$~pp on POPE. The MMStar improvement over uniform LoRA is
supported by a 95\% paired-bootstrap confidence interval of
$[1.4,3.2]$~pp. The MMStar advantage also remains positive relative
to random reweighting ($+2.8$~pp, 95\% CI $[1.6,4.0]$),
accuracy-based mining ($+1.8$~pp, $[0.7,2.9]$), and
ablation-based selection ($+1.3$~pp, $[0.2,2.4]$).
Full statistics are reported in
Appendix~\ref{app:rw_supplementary}.

On MMStar, these matched comparisons indicate that the observed gain
is not explained solely by assigning non-uniform weights, selecting
generically difficult samples, or prioritizing samples that are simply
sensitive to the addition of vision. Instead, the structured
combination of fusion potential, current synergy, and text-unique
reliance provides a more informative ranking signal in the evaluated
reasoning-oriented setting.

\paragraph{The trade-off is selective.}
Relative to uniform LoRA, LoRA-PID decreases MMMU by $0.5$~pp and
PMC-VQA by $0.3$~pp. This trade-off is consistent with the policy's
downweighting of examples exhibiting high text-unique reliance and
suggests that the intervention preferentially benefits settings in
which stronger joint-modality use is useful.


\vspace{-3pt}
\subsection{Where Does PID-Guided Reweighting Help?}
\label{sec:guide_analysis}

\paragraph{Gains concentrate on reasoning-intensive categories.}
The aggregate improvement is not uniform across MMBench categories.
Relative to uniform LoRA, PID-guided reweighting improves
Spatial Reasoning by $+3.5$~pp, Attribute Comparison by $+2.8$~pp,
and Action Recognition by $+2.5$~pp. The gains are smaller for
Object Localization ($+1.1$~pp) and Scene Classification
($+0.5$~pp), and are slightly negative for OCR/Text Recognition and
Commonsense ($-0.3$~pp each).

This pattern is consistent with the intended role of the reweighting
policy: the largest gains occur in categories labeled as
reasoning-oriented in our analysis, whereas the improvements are
smaller or slightly negative for the evaluated perception- and
knowledge-oriented categories. The complete category-level results are
reported in Appendix~\ref{app:rw_supplementary}.

A similar pattern appears on POPE: the largest gain over uniform LoRA
occurs on the Adversarial subset ($+1.9$~pp), where shortcut-prone
co-occurrence cues are particularly challenging
(Appendix~\ref{app:rw_supplementary}).

\vspace{-8pt}
\paragraph{The post-tuning information profile shifts in the targeted
direction.}
LoRA-PID produces the largest increase in held-out MMStar synergy,
from $\Svl=1.20$ after uniform LoRA to $\Svl=1.36$, while
$\Utxt$ decreases from $0.56$ to $0.46$. Consequently, the synergy
share increases from 67.5\% to 73.9\%. The matched alternative
reweighting methods produce smaller shifts in the corresponding
post-tuning PID profile.

Thus, the downstream improvement is accompanied by a change in the
model-induced information profile in the direction targeted by the
reweighting policy. This correspondence does not by itself establish a
causal relationship between increased synergy and performance, but it
provides evidence that the intervention changes the diagnostic profile
it was designed to target. The full decomposition is reported in
Appendix~\ref{app:rw_supplementary}.


\vspace{-3pt}
\subsection{Cross-Model Behavior and Boundary Case}
\label{sec:guide_cross_model}

To examine whether the intervention transfers across different
modality-use profiles, we repeat the comparison on
LLaVA-OneVision-7B and Gemma3-12B.
Table~\ref{tab:rw_cross_model_main} reports the uniform-LoRA and
PID-guided conditions; complete results are provided in
Appendix~\ref{app:rw_cross_model}.

\begin{table}[t]
\centering
\caption{\textbf{Cross-model behavior of PID-guided reweighting.}
Values compare uniform LoRA and PID-guided LoRA.
Post-$\Svl$ is measured on held-out MMStar.}
\label{tab:rw_cross_model_main}
\vspace{-0.4em}
\resizebox{0.94\linewidth}{!}{%
\begin{tabular}{@{}llccc@{}}
\toprule
\textbf{Model}
& \textbf{Condition}
& \textbf{MMBench}
& \textbf{MMStar}
& \textbf{Post-}$\boldsymbol{\Svl}$ \\
\midrule
\multirow{2}{*}{LLaVA-OV-7B}
& LoRA-Uniform
& 85.5{\tiny$\pm$.3}
& 60.0{\tiny$\pm$.4}
& 1.00{\tiny$\pm$.02} \\
& \textbf{LoRA-PID}
& \textbf{86.3}{\tiny$\pm$.3}
& \textbf{61.5}{\tiny$\pm$.4}
& \textbf{1.10}{\tiny$\pm$.03} \\
\midrule
\multirow{2}{*}{Gemma3-12B}
& LoRA-Uniform
& 83.1{\tiny$\pm$.4}
& \textbf{56.2}{\tiny$\pm$.5}
& 0.20{\tiny$\pm$.02} \\
& LoRA-PID
& \textbf{83.3}{\tiny$\pm$.4}
& 54.3{\tiny$\pm$.5}
& 0.21{\tiny$\pm$.02} \\
\bottomrule
\end{tabular}%
}
\vspace{-10pt}
\end{table}

For LLaVA-OneVision-7B, PID-guided reweighting improves MMBench from
85.5 to 86.3 and MMStar from 60.0 to 61.5, while increasing
post-tuning $\Svl$ from 1.00 to 1.10. This reproduces the same
qualitative direction observed for Qwen2.5-VL on a second model with
meaningful joint-modality use.

In contrast, the strongly language-prior-reliant Gemma3-12B does not
show the same benefit. Its MMBench score changes only from 83.1 to
83.3, post-tuning $\Svl$ changes from 0.20 to 0.21, and MMStar
decreases from 56.2 to 54.3. The complete results in
Appendix~\ref{app:rw_cross_model} further show that MMMU decreases
from 59.3 after uniform LoRA to 54.1 after PID-guided reweighting.

This contrast supports a profile-dependent interpretation:
PID-guided reweighting appears most useful when the pretrained model
already exhibits meaningful joint-modality use that can be amplified
through data reweighting. The present experiments do not establish
that the policy can create such fusion capacity when the model's
baseline decisions remain strongly language-prior-reliant.

\begin{mdframed}[
backgroundcolor=gray!5,
linewidth=0.5pt,
innertopmargin=4pt,
innerbottommargin=4pt
]
\textbf{Finding 5.}
PID-guided sample reweighting provides an actionable but
profile-dependent intervention. In the evaluated
joint-modality-reliant models, it improves the reported reasoning and
grounding metrics while shifting post-tuning decisions toward higher
cross-modal synergy; for Qwen2.5-VL-7B, the MMStar gain is additionally
supported by paired-bootstrap confidence intervals. The same benefit is
not observed for the evaluated strongly language-prior-reliant
Gemma3-12B.
\end{mdframed}
\vspace{-1.5pt}
\section{Related Work}
\vspace{-1.5pt}

The shift from vision--language to omni-modal models marks a transition
in generative AI~\cite{luo2025openomni,chen2024omnixr,xu2025qwen2,guo2025m2}.
Existing analyses based on attention, representation alignment, and
modality ablation provide complementary views of multimodal
behavior~\cite{wang2025multishap,basile2025head,fang2026align}.
PID has been used to characterize multimodal information
structure~\cite{fang2026align}, and~\citet{xiu2026a} further
introduced decision-level PID analysis for modality interaction in
LVLMs. However, how omni-modal models integrate multiple sensory
streams under language-conditioned settings remains unclear.
We address this question through \emph{Sensory PID}, which decomposes
video--audio information into unique, redundant, and synergistic
components. Local PID diagnostics further provide signals for VLM
sample reweighting. See Appendix~\ref{app:related_work}.
\vspace{-1.5pt}
\section{Conclusion}
\vspace{-1.5pt}
We introduce \emph{Sensory PID}, a conditional decision-level
decomposition for language-conditioned video--audio interaction in
omni-modal models. It reveals a sensory synergy bottleneck, visual
dominance, visual-first dynamics, and instruction-conditioned sensory
fusion. Local PID diagnostics further provide actionable signals for
VLM sample reweighting, improving reasoning and grounding while
increasing post-tuning cross-modal synergy. Analogous vision--language
analyses serve as reference validation and broadly recover prior
decision-level PID patterns. See limitations in
Appendix~\ref{sec:limitations}.
\section*{Acknowledgments}
We thank the anonymous reviewers and the program committee for their constructive feedback, which helped strengthen the experimental design and clarify the presentation of this work. We also thank our colleagues for helpful discussions during the development of this project. This research is supported by the Ministry of Education, Singapore, under its Academic Research Fund Tier 2 (MOE-T2EP20125-0002) and Tier 1 (RG22/24), National Research Foundation, Singapore under its National Large Language Models Funding Initiative (AISG Award No: AISG-NMLP-2024-001) and NTU Start Up Grant. Any opinions, findings and conclusions or recommendations expressed in this material are those of the author(s) and do not reflect the views of National Research Foundation, Singapore. The computational work for this article was partially performed on resources of the National Supercomputing Centre (NSCC), Singapore (https://www.nscc.sg).
\section*{Impact Statement}This paper presents work whose goal is to advance the field
of machine learning. There are many potential societal
consequences of our work, none of which we feel must be
specifically highlighted here. 

\bibliography{example_paper}
\bibliographystyle{icml2026}

\newpage
\appendix
\onecolumn

\section{Methodology Details and Derivations}
\label{app:method}

This appendix provides methodological details complementary to the
omni-modal Sensory PID formulation in the main text. We first describe
the analogous two-source vision--language PID formulation used for the
reference analyses and PID-guided reweighting experiments. We then
provide the underlying BATCH estimation procedure, the practical
vision--language implementation details, and the derivation of the
BATCH-induced local diagnostic scores used in \S\ref{sec:guide}.


\subsection{Vision--Language PID Reference}
\label{app:vl_pid}

For the vision--language reference setting, let $Y$ denote the model's
prediction over a finite candidate set $\mathcal{C}$ (\eg,
multiple-choice options: \textit{A, B, C, D}), and let $X_v$ and $X_t$
denote the visual and textual source variables, respectively. We
construct an empirical decision distribution by pairing the data
distribution with the model-induced predictive distribution:
\begin{equation}
    P(x_v,x_t,y)
    =
    P_{\mathcal{D}}(x_v,x_t)\,
    p_\theta(y\mid x_v,x_t),
    \label{eq:app_vl_joint}
\end{equation}
with corresponding partial-input distributions induced by
$p_\theta(y\mid x_v)$ and $p_\theta(y\mid x_t)$ through the calibrated
embedding-masking procedure described in
Appendix~\ref{app:batch_estimator}. All information quantities below
are therefore computed with respect to model-induced predictive
distributions rather than ground-truth labels or latent representations.

Partial Information Decomposition (PID)~\citep{williams2010nonnegative}
partitions the decision-relevant information provided by the two
sources into four non-negative components:
\begin{align}
    I(Y;X_v)
    &= \Rvl + \Uvis,
    \label{eq:app_c1}\\
    I(Y;X_t)
    &= \Rvl + \Utxt,
    \label{eq:app_c2}\\
    I(Y;X_v,X_t)
    &= \Rvl + \Uvis + \Utxt + \Svl,
    \label{eq:app_c3}
\end{align}
where $\Uvis$ and $\Utxt$ denote information uniquely attributable to
vision and text, respectively; $\Rvl$ denotes information redundantly
available from either source; and $\Svl$ denotes information available
only from their joint observation.

Following~\citet{bertschinger2014quantifying}, these components are
defined through optimization over the set $\Delta_P$ of joint
distributions $Q(X_v,X_t,Y)$ that preserve the source--target
marginals:
\begin{equation}
    \Delta_P
    =
    \left\{
    Q:
    Q(x_v,y)=P(x_v,y),\;
    Q(x_t,y)=P(x_t,y)
    \right\}.
    \label{eq:app_delta_p}
\end{equation}
Let
\begin{equation}
    M_{\min}
    =
    \min_{Q\in\Delta_P}
    I_Q(Y;X_v,X_t).
    \label{eq:app_mmin}
\end{equation}
The synergistic component is defined as the gap between the true joint
mutual information and the minimum achievable under the preserved
source--target marginals:
\begin{equation}
    \Svl
    =
    I_P(Y;X_v,X_t)
    -
    M_{\min}.
    \label{eq:app_synergy}
\end{equation}
The remaining components follow algebraically:
\begin{align}
    \Utxt
    &=
    M_{\min}
    -
    I_P(Y;X_v),
    \label{eq:app_utxt}\\
    \Uvis
    &=
    M_{\min}
    -
    I_P(Y;X_t),
    \label{eq:app_uvis}\\
    \Rvl
    &=
    I_P(Y;X_v)-\Uvis
    =
    I_P(Y;X_t)-\Utxt.
    \label{eq:app_r}
\end{align}
The four components are non-negative and sum to
$I_P(Y;X_v,X_t)$ by construction.

We use this decomposition as an analogous two-source reference for the
vision--language analyses and for the local PID diagnostics employed
in \S\ref{sec:guide}. Our methodological
focus is instead the conditional Sensory PID formulation
$I(Y;V,A\mid T)$ for language-conditioned video--audio interaction.
The decision-level vision--language formulation follows the framework
used in prior LVLM PID analysis~\citep{xiu2026a} and is not treated as
a methodological contribution of this work.


\subsection{Scalable Estimation via BATCH}
\label{app:batch_estimator}

The PID definitions above require optimization over
$Q\in\Delta_P$, which is intractable for the high-dimensional,
continuous representations produced by modern MLLMs. We use the BATCH
estimator~\citep{liang2023quantifying} as the underlying numerical
estimator. BATCH amortizes the constrained optimization through
learned source encoders and Sinkhorn-normalized couplings.

The derivation below is written using the vision--language notation
$(X_v,X_t)$ because the same two-source formulation underlies the
reference analyses and the local diagnostic scores used for VLM
reweighting in \S\ref{sec:guide}. In the omni-modal analysis, BATCH is
analogously applied to the video and audio sensory sources while the
textual instruction is retained as conditioning context.

\paragraph{Neural parameterization.}
BATCH learns a parametric distribution $\tilde{Q}_\phi$ via two encoder
networks $f_{\phi_1}$ and $f_{\phi_2}$ that map $X_v$ and $X_t$ into a
shared latent space. The unnormalized affinity between sample pairs
within a mini-batch $\mathcal{B}$ of size $m$ is
\begin{equation}
    A_{ij}^{(y)}
    =
    \exp\!\left(
        f_{\phi_1}(x_v^{(i)},y)^\top
        f_{\phi_2}(x_t^{(j)},y)
    \right),
    \qquad
    i,j\in\{1,\ldots,m\}.
    \label{eq:app_affinity}
\end{equation}

\paragraph{Marginal constraints via Sinkhorn--Knopp.}
The PID definition requires $Q\in\Delta_P$. Within a mini-batch, this
corresponds to preserving the required source marginals. BATCH enforces
the constraints by applying the Sinkhorn--Knopp
algorithm~\citep{cuturi2013sinkhorn} to $A^{(y)}$ for each label $y$,
iteratively scaling rows and columns until the resulting matrix
$\tilde{Q}^{(y)}$ is doubly stochastic.

\paragraph{Optimization objective.}
The estimator minimizes an upper bound on
$I_{\tilde{Q}}(Y;X_v,X_t)$ via gradient descent on the encoder
parameters:
\begin{equation}
    \mathcal{L}(\phi)
    =
    \mathbb{E}_{\tilde{Q}}
    \left[
    \log
    \frac{
        \tilde{Q}(x_t\mid x_v,y)\,
        \tilde{Q}(x_v\mid y)
    }{
        \sum_{y'}
        \tilde{Q}(x_t\mid x_v,y')\,
        \tilde{Q}(y'\mid x_v)\,
        \tilde{Q}(x_v)
    }
    \right].
    \label{eq:app_obj}
\end{equation}

\paragraph{PID term recovery.}
Upon convergence, the four PID terms are computed by comparing the
data distribution $P$ against the optimized $\tilde{Q}$:
\begin{align}
    \Svl
    &=
    I_P(Y;X_v,X_t)
    -
    I_{\tilde{Q}}(Y;X_v,X_t),
    \label{eq:app_s_recover}\\
    \Utxt
    &=
    I_{\tilde{Q}}(Y;X_v,X_t)
    -
    I_P(Y;X_v),
    \label{eq:app_u_recover}\\
    \Uvis
    &=
    I_{\tilde{Q}}(Y;X_v,X_t)
    -
    I_P(Y;X_t),
    \label{eq:app_uvis_recover}\\
    \Rvl
    &=
    I_{\tilde{Q}}(Y;X_v,X_t)
    -
    \Utxt
    -
    \Uvis.
    \label{eq:app_r_recover}
\end{align}

\paragraph{Vision--language source representations.}
Token-based MLLMs produce variable-length modality sequences whereas
BATCH requires fixed-dimensional continuous source variables. For the
vision--language reference analysis, we therefore summarize each
modality using mean pooling~\citep{shen2018baseline} after the modality
tokens have been mapped or embedded into the shared multimodal token
space. This yields source representations
$X_v,X_t\in\mathbb{R}^d$. The parameter-free aggregation provides a
global, sample-aligned modality summary for information estimation.
Sensitivity to alternative feature summarization strategies is
reported in Appendix~\ref{app:ablation}.

\paragraph{Calibrated vision--language masking.}
Jointly trained MLLMs do not always expose native unimodal prediction
paths. We therefore approximate the partial-input predictive
distributions through representation corruption, following the broader
use of corruption-based interventions~\citep{meng2022locating}. To
estimate $p_\theta(y\mid x_v)$, we replace the projected text
embeddings while retaining the visual input; conversely, to estimate
$p_\theta(y\mid x_t)$, we replace the visual embeddings while retaining
the textual input. For a masked modality $m$,
\begin{equation}
    \tilde{E}_{m}[\ell]
    \sim
    \mathcal{N}\!\left(
        \mu_m,
        \operatorname{diag}(\sigma_m^2)
    \right),
    \qquad
    \ell=1,\ldots,L_m,
    \label{eq:app_vl_masking}
\end{equation}
where $\mu_m$ and $\sigma_m$ are per-dimension statistics estimated
over the profiling set $\mathcal{D}_{\text{prof}}$. Matching these
empirical statistics controls the perturbation scale while removing
instance-specific information from the masked modality.

\paragraph{Output stabilization.}
For finite-candidate prediction, we apply three stabilization steps
before PID estimation.
First, \emph{valid-option renormalization} restricts predictions to
the candidate set $\mathcal{C}$ through selective softmax over the
corresponding logits.
Second, \emph{confidence gating} prevents low-confidence predictions
from becoming artificially sharp after renormalization: when the total
probability mass assigned to valid options falls below $\tau=0.3$, we
replace the output with a uniform distribution over $\mathcal{C}$.
Third, rather than constructing the target marginal from hard argmax
labels, we preserve predictive uncertainty through
\emph{soft aggregation}:
\begin{equation}
    \hat{p}(y)
    =
    \frac{1}{|\mathcal{D}_{\text{prof}}|}
    \sum_{j\in\mathcal{D}_{\text{prof}}}
    p_\theta
    \left(
        y\mid x_v^{(j)},x_t^{(j)}
    \right),
    \label{eq:app_vl_marginal}
\end{equation}
and keep this marginal fixed during BATCH optimization.

\paragraph{Estimator provenance.}
BATCH, mean pooling, embedding-space corruption, confidence
regularization, and soft aggregation are estimation machinery rather
than methodological contributions of this work. Closely related
combinations of these choices have been reported in prior
decision-level LVLM analysis~\citep{xiu2026a}. The methodological
contribution emphasized in the main text is instead the conditional
Sensory PID formulation and its instantiation for
language-conditioned video--audio interaction.


\subsection{BATCH-Induced Local Diagnostic Scores}
\label{sec:method_local}

For the PID-guided reweighting application in \S\ref{sec:guide}, we
further retain the sample-level additive contributions underlying the
aggregate vision--language PID estimates. The four mutual-information
quantities underlying the decomposition---
$I_P(Y;X_v)$,
$I_P(Y;X_t)$,
$I_P(Y;X_v,X_t)$, and
$I_{\tilde{Q}}(Y;X_v,X_t)$---
are implemented as expectations over samples. Retaining their
sample-level terms before averaging yields the BATCH-induced local
diagnostic contributions
\begin{align}
    s_i
    &=
    \text{mi}_{P}^{(i)}(Y;X_v,X_t)
    -
    \text{mi}_{\tilde{Q}}^{(i)}(Y;X_v,X_t),
    \label{eq:local_s}\\
    u_{\text{txt},i}
    &=
    \text{mi}_{\tilde{Q}}^{(i)}(Y;X_v,X_t)
    -
    \text{mi}_{P}^{(i)}(Y;X_v),
    \label{eq:local_utxt}\\
    u_{\text{vis},i}
    &=
    \text{mi}_{\tilde{Q}}^{(i)}(Y;X_v,X_t)
    -
    \text{mi}_{P}^{(i)}(Y;X_t),
    \label{eq:local_uvis}\\
    r_i
    &=
    \text{mi}_{P}^{(i)}(Y;X_v)
    +
    \text{mi}_{P}^{(i)}(Y;X_t)
    -
    \text{mi}_{\tilde{Q}}^{(i)}(Y;X_v,X_t).
    \label{eq:local_r}
\end{align}
For example,
\begin{equation}
    \text{mi}_{P}^{(i)}(Y;X_v)
    =
    \sum_y
    p_\theta(y\mid x_v^{(i)})
    \log
    \frac{
        p_\theta(y\mid x_v^{(i)})
    }{
        \hat{p}(y)
    },
    \label{eq:app_local_example}
\end{equation}
is the local MI contribution from the vision-only predictive
distribution, while
$\text{mi}_{\tilde{Q}}^{(i)}$ additionally involves the learned
Sinkhorn coupling.

Within each batch context, the aggregate PID estimate is recovered by
averaging these local contributions:
$\hat{S}=|\mathcal{B}|^{-1}\sum_i s_i$, and analogously for
$\hat{U}_{\text{txt}}$, $\hat{U}_{\text{vis}}$, and $\hat{R}$.

These local scores are \emph{not} axiomatic pointwise PID atoms---a
formal theory of pointwise PID remains an open
problem~\citep{liang2023quantifying}. We use them exclusively as
BATCH-induced additive \emph{ranking signals}. Because the Sinkhorn
coupling depends on batch composition, we stabilize each sample's
score by averaging over $K=50$ random batch draws with batch size 256.

To construct the reweighting signals used in \S\ref{sec:guide}, we
apply non-negative clipping to the local contributions and define the
per-sample information mass
\begin{equation}
    I_i^{+}
    =
    [s_i]_+
    +
    [u_{\text{vis},i}]_+
    +
    [u_{\text{txt},i}]_+
    +
    [r_i]_+,
    \qquad
    [z]_+=\max(z,0).
    \label{eq:app_information_mass}
\end{equation}
The \emph{synergy ratio} and \emph{shortcut score} are then
\begin{equation}
    \SR_i
    =
    \frac{[s_i]_+}{I_i^{+}+\epsilon},
    \qquad
    \SC_i
    =
    \frac{[u_{\text{txt},i}]_+}{I_i^{+}+\epsilon},
    \label{eq:sr_sc_app}
\end{equation}
where $\epsilon=10^{-8}$. $\SR_i$ serves as a proxy for the relative
synergy usage of sample $i$, while $\SC_i$ captures its relative
text-unique reliance. These ratios convert aggregate interaction
diagnostics into sample-level ranking signals used by the
PID-guided reweighting policy in \S\ref{sec:guide}.


\subsection{BATCH-Induced Local Diagnostic Scores: Full Derivation}
\label{app:local_scores}

Appendix~\ref{sec:method_local} introduces the BATCH-induced
sample-level diagnostic scores used for reweighting in
\S\ref{sec:guide}. Here we provide the complete derivation and the
stability checks used in our experiments.

\paragraph{Per-sample MI terms.}
The four MI quantities underlying the PID---
$I_P(Y;X_v)$,
$I_P(Y;X_t)$,
$I_P(Y;X_v,X_t)$, and
$I_{\tilde{Q}}(Y;X_v,X_t)$---
are each implemented as sample-level expectations. The partial-input
terms use the calibrated vision--language masking procedure described
in Appendix~\ref{app:batch_estimator}:
\begin{align}
    \text{mi}_{P}^{(i)}(Y;X_v)
    &=
    \sum_{y\in\mathcal{C}}
    p_\theta(y\mid x_v^{(i)})
    \log
    \frac{
        p_\theta(y\mid x_v^{(i)})
    }{
        \hat{p}(y)
    },
    \label{eq:app_mi_xv}\\
    \text{mi}_{P}^{(i)}(Y;X_t)
    &=
    \sum_{y\in\mathcal{C}}
    p_\theta(y\mid x_t^{(i)})
    \log
    \frac{
        p_\theta(y\mid x_t^{(i)})
    }{
        \hat{p}(y)
    },
    \label{eq:app_mi_xt}\\
    \text{mi}_{P}^{(i)}(Y;X_v,X_t)
    &=
    \sum_{y\in\mathcal{C}}
    p_\theta(y\mid x_v^{(i)},x_t^{(i)})
    \log
    \frac{
        p_\theta(y\mid x_v^{(i)},x_t^{(i)})
    }{
        \hat{p}(y)
    }.
    \label{eq:app_mi_joint}
\end{align}
Here
\begin{equation}
    \hat{p}(y)
    =
    \frac{1}{|\mathcal{D}_{\text{prof}}|}
    \sum_j
    p_\theta
    \left(
        y\mid x_v^{(j)},x_t^{(j)}
    \right)
    \label{eq:app_marginal_repeat}
\end{equation}
is the marginal label distribution estimated once from the profiling
set.

\paragraph{Row-anchored coupling MI.}
The fourth term additionally involves the learned Sinkhorn coupling.
Fixing the $X_v$ index at sample $i$ and marginalizing over all $X_t$
samples in the batch gives
\begin{equation}
\begin{aligned}
    \text{mi}_{\tilde{Q}}^{(i)}(Y;X_v,X_t)
    =
    \sum_{j\in\mathcal{B}}
    \sum_y
    &\,
    p_\theta(y\mid x_v^{(i)})
    \tilde{q}
    \left(
        x_t^{(j)}
        \mid
        x_v^{(i)},y
    \right)
    \\
    &\times
    \log
    \frac{
        p_\theta(y\mid x_v^{(i)})
        \tilde{q}
        \left(
            x_t^{(j)}
            \mid
            x_v^{(i)},y
        \right)
    }{
        \hat{p}(y)
        \tilde{q}
        \left(
            x_t^{(j)}
            \mid
            x_v^{(i)}
        \right)
    } .
\end{aligned}
\label{eq:app_mi_q}
\end{equation}
Here
$\tilde{q}(x_t^{(j)}\mid x_v^{(i)},y)$ is obtained by normalizing the
$i$-th row of the Sinkhorn matrix associated with label $y$, and
\begin{equation}
    \tilde{q}
    \left(
        x_t^{(j)}
        \mid
        x_v^{(i)}
    \right)
    =
    \sum_{y'}
    \tilde{q}
    \left(
        x_t^{(j)}
        \mid
        x_v^{(i)},y'
    \right)
    p_\theta
    \left(
        y'\mid x_v^{(i)}
    \right).
    \label{eq:app_q_marginal}
\end{equation}

\paragraph{Local PID scores.}
Substituting the sample-level MI quantities above into the aggregate
PID identities gives
\begin{align}
    s_i(\mathcal{B})
    &=
    \text{mi}_{P}^{(i)}(Y;X_v,X_t)
    -
    \text{mi}_{\tilde{Q}}^{(i)}(Y;X_v,X_t),
    \label{eq:app_local_s_full}\\
    u_{\text{txt},i}(\mathcal{B})
    &=
    \text{mi}_{\tilde{Q}}^{(i)}(Y;X_v,X_t)
    -
    \text{mi}_{P}^{(i)}(Y;X_v),
    \label{eq:app_local_utxt_full}\\
    u_{\text{vis},i}(\mathcal{B})
    &=
    \text{mi}_{\tilde{Q}}^{(i)}(Y;X_v,X_t)
    -
    \text{mi}_{P}^{(i)}(Y;X_t),
    \label{eq:app_local_uvis_full}\\
    r_i(\mathcal{B})
    &=
    \text{mi}_{P}^{(i)}(Y;X_v)
    +
    \text{mi}_{P}^{(i)}(Y;X_t)
    -
    \text{mi}_{\tilde{Q}}^{(i)}(Y;X_v,X_t).
    \label{eq:app_local_r_full}
\end{align}
The batch-level PID estimates recover by averaging these terms:
\begin{equation}
    \hat{S}
    =
    \frac{1}{|\mathcal{B}|}
    \sum_i
    s_i(\mathcal{B}),
    \label{eq:app_batch_recovery}
\end{equation}
and analogously for
$\hat{U}_{\text{txt}}$,
$\hat{U}_{\text{vis}}$, and $\hat{R}$.

\paragraph{Row vs.\ column anchoring.}
One can alternatively construct the local coupling contribution by
fixing $X_t$ and marginalizing over $X_v$, yielding a column-anchored
score. We use row-anchored scores by default and evaluate whether this
choice materially changes sample ranking. Across the 3{,}000 profiled
samples, the Spearman correlation between row- and column-anchored
$\bar{s}_i$ scores is
$\rho=0.87$ (95\% CI $[0.83,0.91]$), indicating broadly consistent
rankings.

\paragraph{Multi-context averaging.}
Because the Sinkhorn coupling depends on the composition of the
mini-batch, an individual sample's local score can vary across batch
contexts. We therefore average each score over $K=50$ random batch
draws with batch size 256:
\begin{equation}
    \bar{s}_i
    =
    \frac{1}{K}
    \sum_{k=1}^{K}
    s_i(\mathcal{B}_k),
    \qquad
    \bar{u}_{\text{txt},i}
    =
    \frac{1}{K}
    \sum_{k=1}^{K}
    u_{\text{txt},i}(\mathcal{B}_k).
    \label{eq:app_avg}
\end{equation}
The same averaging procedure is applied to the remaining local
diagnostic contributions before constructing the reweighting scores
used in \S\ref{sec:guide}.

\section{Extended Related Work}
\label{app:related_work}

\textbf{Multimodal Large Language Models and Evaluation.}
The progression from vision--language models (VLMs) to omni-modal systems capable of processing interleaved video, audio, and text marks a major shift in generative AI~\cite{fu2025vita,xu2025qwen2}. Early systems typically relied on separate modality encoders connected to a language-model backbone, while recent native multimodal architectures increasingly fuse sensory inputs within a shared token space~\cite{zhang2026can,dai2023instructblip,liu2024improved,luo2025openomni}. Despite strong benchmark performance, evaluation remains largely outcome-oriented, relying on accuracy-based leaderboards~\cite{zheng2023judging} or human preference ratings~\cite{zhang2025omnieval}. Such metrics are useful for ranking models, but they provide limited insight into how different modalities contribute to the final decision. Our work shifts the focus from \textit{what} a model predicts to \textit{how} sensory and linguistic information is used at the decision level.

\textbf{Interpretability and Modality Interaction Analysis.} Recent work on LLM interpretability has developed a broad set of tools for explaining model behavior beyond final predictions, including attribution, probing, mechanistic analysis, and representation-level inspection~\cite{ma2026unimodalshortcutsmllmscrossmodal,zhang2025evaluating,zhang2026instruction}. Understanding modality interaction is essential for reliable multimodal deployment. Existing approaches can be broadly grouped into attention analysis, representation alignment, and modality ablation. Attention-based methods visualize token- or head-level associations~\cite{jain2019attention,chefer2021transformer,basile2025head}, but attention weights do not necessarily reflect functional contribution. Representation analyses measure geometric or statistical alignment between modality embeddings~\cite{morcos2018insights,nguyen2020wide,radford2021learning}; however, high alignment does not imply that the aligned information is used in the final prediction. Ablation studies quantify behavioral sensitivity under modality removal or perturbation~\cite{jiao2024comprehensive,tong2024cambrian,thrush2022winoground}, but they conflate qualitatively different sources of dependence: a performance drop may arise from modality-unique information, redundant evidence, or genuinely synergistic integration. In contrast, our PID framework explicitly separates unique, redundant, and synergistic decision information, enabling a finer-grained characterization of modality use than accuracy, alignment, or ablation alone.

\textbf{Information-Theoretic Analysis in Deep Learning.}
Information theory provides a general framework for quantifying uncertainty, dependence, and information flow, and has been widely applied across communication, neuroscience, complex systems, and machine learning~\cite{du2024bottleneck,kim2026does,huang2026llm2clip}. The Information Bottleneck principle~\cite{tishby2015deep} has been widely used to study compression and prediction in neural networks, while Partial Information Decomposition (PID)~\cite{williams2010nonnegative} separates the information that multiple sources provide about a target into unique, redundant, and synergistic components. Recent estimators and scalable approximations~\cite{faes2016information,kolchinsky2022novel,wibral2017partial,bertschinger2014quantifying,liang2023quantifying} have made PID increasingly applicable to high-dimensional systems. In multimodal learning, PID has been used to characterize
redundancy--uniqueness--synergy structure across modality pairs and to
inform representation alignment~\citep{fang2026align}. More recently,
~\citet{xiu2026a} study PID over model-induced predictive
distributions in LVLMs, enabling decision-level analysis of
vision--language interaction; related information-decomposition
perspectives have also been explored in diffusion models
\citep{dewan2024diffusion}.

Our work focuses on a complementary setting in which language specifies
the task and video and audio provide sensory evidence. We introduce
\emph{Sensory PID}, which conditions on language and decomposes
$I(Y;V,A\mid T)$ into video-unique, audio-unique, redundant, and
audiovisual-synergistic information. This formulation supports analyses
specific to omni-modal sensory interaction, including modality-specific
perturbations, visual-first layer-wise dynamics, and
instruction-conditioned sensory fusion. Separately, we use
BATCH-induced local PID diagnostics as ranking signals for VLM sample
reweighting, connecting information-theoretic diagnosis with a
proof-of-concept training intervention.

\section{Limitations and Future Directions}
\label{sec:limitations}

While PID and Sensory PID provide a principled framework for analyzing multimodal interaction, we acknowledge several limitations of the current study.

\textbf{Constrained Decoding Space.}
First, our estimation framework operates over a discrete decision space, primarily multiple-choice QA. This setting makes the model's predictive distribution tractable and enables stable PID estimation, but it does not fully capture open-ended generation. Extending PID to free-form responses remains challenging because mutual information must be estimated over high-dimensional, sequential output distributions. Future work will require more efficient estimators for sequence-level or semantic-level decision spaces.

\textbf{Approximation of Unimodal Conditionals.}
Second, our training-free analysis relies on calibrated embedding masking to approximate unimodal conditionals such as $P(Y \mid V,T)$ or $P(Y \mid A,T)$. Although moment-matched noise helps preserve distributional stability, masking remains a proxy for native unimodal inference. The resulting hidden states may differ from those produced by models explicitly trained or prompted for partial-modality inputs, potentially introducing estimation bias in unique information terms such as $U_{\mathrm{vis}}$ and $U_{\mathrm{aud}}$.

\textbf{Statistical Rather than Causal Interpretation.}
Third, PID measures statistical dependence in model-induced predictive distributions; it is not a causal calculus. While PID profiles predict sensitivity to modality-level interventions, the PID terms themselves do not identify the internal circuits, attention heads, or neurons responsible for the observed behavior. Thus, high synergy is consistent with functional cross-modal integration, but it does not by itself reveal the mechanistic implementation of that integration.

\textbf{Scope of PID-Guided Reweighting.}
Finally, our PID-guided reweighting experiment is intended as a proof-of-concept rather than a complete training recipe. We use lightweight LoRA fine-tuning and a limited reweighting policy to test whether local diagnostic scores contain actionable signal. Scaling this idea to larger corpora, stronger adaptation methods, and open-ended instruction tuning remains an important direction for future work.

\paragraph{Future Directions.}
Building on these limitations, several directions are particularly promising.
\begin{itemize}
    \item \textbf{Extending PID beyond multiple-choice decisions.}
    Future work could generalize the estimator to continuous or autoregressive output spaces, enabling analysis of chat-style assistants, dense captioning, and long-form multimodal generation.

    \item \textbf{Using PID as a training-time diagnostic.}
    PID terms could serve as monitoring signals during multimodal training or scaling analysis, helping identify when models rely on language shortcuts, redundant evidence, or genuine cross-modal integration. More speculative extensions could explore PID-inspired auxiliary objectives that encourage fusion-sensitive behavior, while carefully avoiding over-optimization of any single information term.

    \item \textbf{PID-guided benchmark and data construction.}
    PID analysis can help curate evaluation or training sets that better isolate cross-modal reasoning. For example, samples with high text-unique reliance may reveal language-prior shortcuts, whereas high-synergy samples can support more targeted evaluation of multimodal fusion. This may lead to more informative benchmarks than those defined solely by surface-level task categories.
\end{itemize}

\section{Experimental Setup Details}
\label{app:setup}

\subsection{Prompt Templates}
\label{app:prompts}

To ensure fair evaluation and reproducible results, we employ a unified prompt structure across all benchmarks. Table~\ref{tab:prompt_templates} details the specific templates used for standard 4-option multiple-choice tasks (e.g., MMBench, MMMU) and binary 2-option tasks (e.g., specific subsets of MUSIC-AVQA or Boolean questions).

\begin{table}[h]
    \centering
    \caption{Prompt templates used for zero-shot evaluation. The content within \texttt{< >} is replaced by the specific instance data during inference.}
    \label{tab:prompt_templates}
    \vspace{0.2cm}
    \renewcommand{\arraystretch}{1.5}
    \begin{tabular}{p{0.25\textwidth} | p{0.7\textwidth}}
        \toprule
        \textbf{Task Type} & \textbf{Prompt Template} \\
        \midrule
        \textbf{4-Option Questions} \newline (Standard MCQ) & 
        Question: \texttt{<Question Body>} \newline
        Options: \newline
        (A) \texttt{<Option A>} \newline
        (B) \texttt{<Option B>} \newline
        (C) \texttt{<Option C>} \newline
        (D) \texttt{<Option D>} \newline
        Instruction: Please analyze and select the most appropriate answer. Output only the single letter corresponding to the correct option above (e.g., A, B, C, or D). Do not provide any explanation. \\
        \midrule
        \textbf{2-Option Questions} \newline (Binary/True-False) & 
        Question: \texttt{<Question Body>} \newline
        Options: \newline
        (A) \texttt{<Option A>} \newline
        (B) \texttt{<Option B>} \newline
        Instruction: Please analyze and select the most appropriate answer. Output only the single letter corresponding to the correct option above (e.g., A or B). Do not provide any explanation. \\
        \bottomrule
    \end{tabular}
\end{table}

\subsection{Inference Settings}
For all evaluations, we use deterministic decoding strategies to minimize variance. Specifically, we set the temperature to $0$ and top-p to $1.0$. The maximum output token length is restricted to ensure the model adheres to the single-letter constraint.

\subsection{Hyperparameters and Training Details}
\label{app:hyperparameters}

To ensure the full reproducibility of our PID or sensory PID estimation results, we provide the detailed hyperparameter configurations used for the BATCH estimator. Following the guidelines established in previous work \cite{liang2023quantifying}, we employed a lightweight Multi-Layer Perceptron (MLP) parameterization to approximate the marginal-matching joint distributions.

We adhered to a fixed configuration across all experiments (Vision-Language and Omni-modal tasks) to ensure consistency. Table~\ref{tab:batch_hyperparameters} summarizes the specific settings. While the estimator is generally robust to hyperparameter variations, we report the exact values used to generate the main results in this paper.

\begin{table}[h]
    \centering
    \caption{Hyperparameters and configuration details for the BATCH Estimator.}
    \label{tab:batch_hyperparameters}
    \vspace{0.2cm}
    \renewcommand{\arraystretch}{1.3} 
    \begin{tabular}{l|c}
        \toprule
        \textbf{Hyperparameter} & \textbf{Value / Configuration} \\
        \midrule
        \multicolumn{2}{c}{\textit{Optimization}} \\
        \midrule
        Optimizer & Adam \\
        Learning Rate Set & $1 \times 10^{-3}$ and $1 \times 10^{-4}$ \\
        Training Epochs Set & 10, 15, 20 \\
        Batch Size Set (Training \& Inference) & 64/128/256 \\
        \midrule
        \multicolumn{2}{c}{\textit{Network Architecture}} \\
        \midrule
        Encoder Network & 3-layer MLP \\
        Hidden Dimension & 32 \\
        Activation Function & ReLU \\
        Initialization & Standard Xavier \\
        \bottomrule
    \end{tabular}
\end{table}
\section{Dataset Details and Taxonomy}
\label{app:datasets}

In this section, we provide detailed specifications of the benchmarks used in our analysis. For the vision--language benchmarks, we organize datasets according to the \textbf{interaction regimes identified by our PID analysis} in the main text. These categories are analytical summaries of the observed model--benchmark profiles, rather than labels assigned before applying PID. We also describe the subsets constructed for the omni-modal analysis.

\subsection{Vision--Language Benchmarks}

\paragraph{Regime I: Synergy-Dominant Reasoning and Grounding Benchmarks.}
These benchmarks exhibit relatively high cross-modal synergy ($S_{\mathrm{vl}}$) in our experiments, suggesting that successful model decisions tend to depend on joint use of visual evidence and textual queries.

\begin{itemize}
    \item \textbf{MMBench}~\cite{liu2024mmbench}: 
    MMBench is a systematically designed benchmark for evaluating LVLMs across approximately 20 fine-grained ability dimensions, ranging from basic object perception and localization to relation reasoning and attribute understanding. Its hierarchical taxonomy and quality-control pipeline make it a broad testbed for general multimodal capability. In our PID analysis, MMBench tends to elicit synergy-dominant profiles across fusion-capable model families, indicating that visual and textual information are often jointly used in the final decision.

    \item \textbf{MMStar}~\cite{chen2024we}: 
    MMStar is a vision-indispensable benchmark designed to reduce multimodal leakage and filter out samples that can be solved from language priors alone. It contains carefully validated multiple-choice questions where visual information is expected to be necessary for correct answering. Consistent with this design goal, our PID results show elevated $S_{\mathrm{vl}}$ on MMStar, making it a representative benchmark for studying cross-modal integration in vision--language models.

    \item \textbf{POPE}~\cite{li2023evaluating}: 
    POPE is an object hallucination benchmark that probes whether LVLMs correctly judge the presence or absence of objects in an image through binary yes/no questions. Because the task directly tests object-level grounding, successful predictions often require matching textual object queries with visual evidence rather than relying solely on co-occurrence priors. In our analysis, POPE exhibits a grounding-sensitive, synergy-dominant profile, separating it from other hallucination-oriented benchmarks whose decisions may depend more strongly on language priors.
\end{itemize}

\paragraph{Regime II: Text-Unique / Prior-Dominant Knowledge Benchmarks.}
These benchmarks exhibit relatively high language-unique information ($U_{\mathrm{txt}}$) in our experiments. Although they contain visual inputs, the evaluated models often rely strongly on textual context, answer priors, or domain knowledge when forming decisions.

\begin{itemize}
    \item \textbf{MMMU (Massive Multi-discipline Multimodal Understanding)}~\cite{yue2024mmmu}: 
    MMMU is a large-scale benchmark for college-level multidisciplinary multimodal understanding. It spans multiple disciplines and includes heterogeneous image types such as charts, diagrams, tables, chemical structures, and scientific illustrations. While visual interpretation is important for many questions, the benchmark also requires substantial domain knowledge and reasoning over specialized concepts. In our PID decomposition, MMMU tends to exhibit stronger $U_{\mathrm{txt}}$, suggesting that model decisions are often influenced by language-side domain knowledge and answer priors.

    \item \textbf{PMC-VQA}~\cite{zhang2023pmc}: 
    PMC-VQA is a medical visual question-answering dataset derived from biomedical and clinical imagery, including modalities such as X-rays, MRIs, pathology images, and other medical figures. The questions often require specialized medical knowledge for interpretation and answer selection. Similar to MMMU, our analysis places PMC-VQA in a text-unique / prior-dominant regime, where language-side medical knowledge contributes substantially to the model's predictive distribution.

    \item \textbf{Reefknot}~\cite{zheng2025reefknot}: 
    Reefknot evaluates relational hallucination in LVLMs through multiple-choice questions involving spatial or semantic relationships among visual entities. Although it is related to hallucination evaluation, its interaction profile differs from POPE in our PID analysis. Whereas POPE focuses on object-level existence grounding, Reefknot often involves relational patterns that can be partially supported by textual priors or common visual-language associations. Our results show that Reefknot elicits stronger $U_{\mathrm{txt}}$ than POPE, illustrating that benchmarks with similar surface-level labels can induce distinct modality-use profiles.
\end{itemize}

\subsection{Omni-Modal Benchmark}

For the tri-modal analysis ($V, A, T$), we utilize a specific dataset capable of isolating audio-visual dependencies.

\begin{itemize}
    \item \textbf{MUSIC-AVQA} \cite{li2022learning}: A large-scale audio-visual dataset focused on music performance videos. The original dataset contains over 45K QA pairs encompassing spatial-temporal relations and acoustic comparisons.
    
    \textbf{Our Evaluation Split:} To rigorously test the ``Sensory Synergy Bottleneck,'' we do not use the raw random split. Instead, we construct a balanced evaluation set consisting of three distinct subsets (1800 samples each) based on the dataset's metadata:
    \begin{enumerate}
        \item \textbf{Audio-Focused:} Questions solvable primarily via auditory cues (e.g., identifying instruments by sound).
        \item \textbf{Visual-Focused:} Questions relying on visual objects or actions (e.g., identifying visible instruments).
        \item \textbf{AV-Fusion:} Questions explicitly requiring joint reasoning (e.g., ``Is the violin playing the higher pitch?''). This subset serves as our critical testbed for identifying genuine sensory synergy ($S_{av}$).
    \end{enumerate}
\end{itemize}


\section{Vision--Language Reference Analysis}
\label{app:full_vl}

This appendix reports the complete vision--language reference analyses
summarized briefly in the main text. These experiments apply the
analogous decision-level two-source PID formulation described in
Appendix~\ref{app:vl_pid} to 20 vision--language models across six
benchmarks. Their purpose is to validate the implementation against
previously reported LVLM PID behavior and to provide the
vision--language diagnostics used by the reweighting study, rather than
to constitute the primary empirical contribution of the paper.


\subsection{Benchmark-Level Interaction Profiles}
\label{sec:diagnose}

\begin{figure*}[t]
\centering
\includegraphics[width=\textwidth]{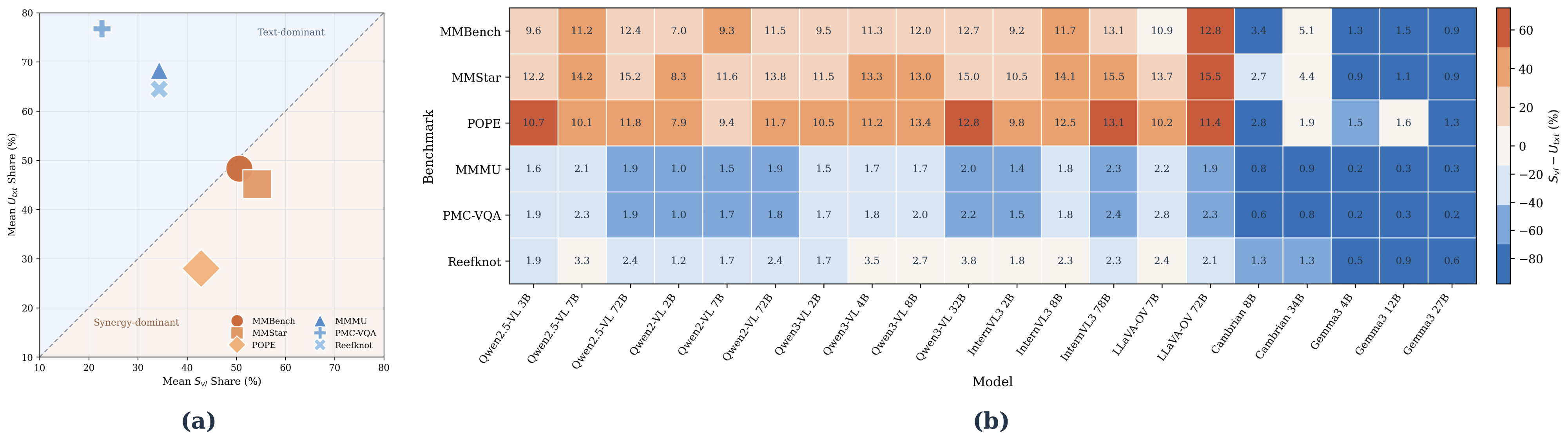}
\caption{\textbf{PID-derived modality-use analysis across 20 models
and 6 vision--language benchmarks.}
\textbf{(a)} Benchmark landscape: each marker plots one benchmark's
mean $\Svl$ share versus mean $\Utxt$ share, averaged across all
20 models. The dashed diagonal marks $\Svl=\Utxt$; points above the
diagonal exhibit relatively higher synergy, whereas points below
exhibit relatively higher text-unique information.
\textbf{(b)} Model--benchmark interaction profiles. Cell color denotes
the profile gap ($\Svl$ share $-$ $\Utxt$ share, pp), with positive
values indicating relatively stronger synergy and negative values
indicating relatively stronger text-unique reliance. Cell values report
$\Delta_{\text{vision}}=\text{Acc}(V,T)-\text{Acc}(T)$ (pp), where
larger values indicate stronger behavioral dependence on vision.}
\label{fig:combined}
\end{figure*}

Figure~\ref{fig:combined}(a) plots each benchmark's mean $\Svl$ share
against its mean $\Utxt$ share, averaged across all 20 models.
Consistent with the task-level separation reported by Xiu
et al.~\cite{xiu2026a}, the six-benchmark evaluation exhibits a similar
division between relatively synergy-dominant and
text-unique-dominant profiles.

MMBench, MMStar, and POPE fall on the $\Svl>\Utxt$ side of the
diagonal, indicating relatively stronger cross-modal synergy, whereas
MMMU, PMC-VQA, and Reefknot fall on the $\Utxt>\Svl$ side, with
$\Utxt$ shares exceeding 60\%. We also recover the previously reported
contrast between POPE and Reefknot~\cite{xiu2026a}: despite both
targeting hallucination-related behavior, POPE exhibits a more
grounding-sensitive, synergy-heavy profile, whereas Reefknot shows
stronger text-unique reliance. The additional MMStar and MMMU
benchmarks follow the same broader separation.

\paragraph{Reference observation.}
Across the expanded benchmark set, the decision-level PID profiles
broadly corroborate previously reported task-dependent modality-use
regimes~\cite{xiu2026a}, while extending the comparison to MMStar and
MMMU.


\subsection{Model--Benchmark Interaction Profiles}
\label{sec:interaction}

We next examine whether family-level modality-use tendencies persist
across benchmarks and model scales. Consistent with prior LVLM PID
analysis~\cite{xiu2026a}, Figure~\ref{fig:combined}(b) shows recurring
differences across model families. Qwen2/2.5-VL, Qwen3-VL, InternVL3,
and LLaVA-OneVision generally exhibit more
\emph{joint-modality-reliant} profiles, with higher $\Svl$ and larger
$\Delta_{\text{vision}}$ on synergy-dominant benchmarks, whereas
Gemma3 and Cambrian tend toward stronger
\emph{language-prior-reliant} profiles with higher $\Utxt$. These
tendencies vary continuously across benchmarks rather than forming a
strict dichotomy.

Scaling generally preserves these family-level tendencies. Notably,
Qwen3-VL maintains a similar profile from 2B to 32B, while Gemma3
remains comparatively text-unique-heavy from 4B to 27B. Variation
across benchmarks further indicates that these profiles are best
interpreted as model--benchmark interactions rather than intrinsic
properties of models alone.

\paragraph{Reference observation.}
Family-level modality-use tendencies broadly corroborate prior LVLM PID
findings~\cite{xiu2026a} and persist in the updated evaluation,
including Qwen3-VL and the additional MMStar and MMMU benchmarks.


\subsection{Behavioral and Intervention Validity}
\label{sec:predict}

We examine whether the reference PID profiles are associated with
observable model behavior. For each model--benchmark pair, we define
\begin{equation}
    \Delta_{\text{vision}}
    =
    \operatorname{Acc}(V,T)
    -
    \operatorname{Acc}(T),
\end{equation}
where larger values indicate stronger behavioral dependence on vision.

\begin{table*}[t]
\centering
\captionsetup{font=footnotesize}
\caption{Spearman $\rho$ between PID quantities and two behavioral measures across 20 models. Each cell reports $\rho(\cdot,\Delta_{\text{vision}})$ / $\rho(\cdot,\text{Acc})$. $I$: total MI; CoI: co-information ($\Rvl{-}\Svl$). ${}^{\ddagger}\!p{<}.001$; ${}^{\dagger}\!p{<}.01$; ${}^{*}\!p{<}.05$; ${}^{\circ}\!p{<}.10$. Bold: $|\rho|{\geq}0.68$.}
\label{tab:spearman_full}
\tiny
\setlength{\tabcolsep}{3pt}
\renewcommand{\arraystretch}{1.0}
\resizebox{\textwidth}{!}{%
\begin{tabular}{@{}ll cccc cc@{}}
\toprule
\textbf{Regime} & \textbf{Benchmark}
& $\Svl$ & $\Utxt$ & $\Uvis$ & $\Rvl$ & $I(V,\!T;\!Y)$ & CoI \\
\midrule
\multirow{3}{*}{\makecell[c]{Synergy-\\driven}}
& MMBench
& $\mathbf{0.840}^{\ddagger} / \mathbf{0.752}^{\ddagger}$
& $-0.582^{\dagger} / 0.198$
& $0.068 / 0.058$
& $0.098 / 0.088$
& $-0.118 / 0.522^{*}$
& $\mathbf{-0.822}^{\ddagger} / \mathbf{-0.682}^{\ddagger}$ \\
& MMStar
& $\mathbf{0.862}^{\ddagger} / \mathbf{0.762}^{\ddagger}$
& $-0.548^{*} / 0.148$
& $0.042 / 0.032$
& $0.062 / 0.052$
& $-0.082 / 0.478^{*}$
& $\mathbf{-0.850}^{\ddagger} / \mathbf{-0.718}^{\ddagger}$ \\
& POPE
& $\mathbf{0.798}^{\ddagger} / \mathbf{0.718}^{\ddagger}$
& $-0.502^{*} / 0.098$
& $0.078 / 0.068$
& $0.348 / 0.298$
& $0.052 / 0.418^{\circ}$
& $\mathbf{-0.722}^{\ddagger} / -0.622^{\dagger}$ \\
\midrule
\multirow{3}{*}{\makecell[c]{Prior-\\dominant}}
& MMMU
& $0.318 / 0.391$
& $-0.198 / 0.422^{*}$
& $0.038 / 0.058$
& $0.048 / 0.042$
& $-0.052 / 0.548^{*}$
& $-0.298 / -0.242$ \\
& PMC-VQA
& $0.382^{\circ} / 0.398^{\circ}$
& $-0.178 / 0.356^{*}$
& $0.052 / 0.042$
& $0.032 / 0.028$
& $-0.078 / 0.593^{\dagger}$
& $-0.348 / -0.317$ \\
& Reefknot
& $0.418^{\circ} / 0.348$
& $-0.278 / 0.378^{*}$
& $0.068 / 0.072$
& $0.058 / 0.048$
& $0.022 / 0.552^{*}$
& $-0.382^{\circ} / -0.278$ \\
\bottomrule
\end{tabular}%
}
\end{table*} 

Table~\ref{tab:spearman_full} reports Spearman correlations between
the PID quantities, accuracy, and $\Delta_{\text{vision}}$ across the
20 models for each benchmark.

\paragraph{PID components track regime-dependent behavior.}
Consistent with prior observations~\cite{xiu2026a}, on MMBench,
MMStar, and POPE, $\Svl$ is strongly associated with both accuracy and
vision-removal sensitivity:
\[
\rho(\Svl,\Delta_{\text{vision}})\geq0.798,
\qquad
\rho(\Svl,\operatorname{Acc})\geq0.718
\quad (p<0.001).
\]
By contrast, $\Utxt$ is negatively associated with
$\Delta_{\text{vision}}$ ($\rho\leq-0.502$, $p<0.05$). On MMMU,
PMC-VQA, and Reefknot, synergy is less dominant and text-unique
information becomes relatively more prominent, while vision-removal
effects remain small ($\Delta_{\text{vision}}<3.5$~pp). MMStar and
MMMU extend these regime-dependent patterns to additional benchmarks.

\paragraph{Total MI captures different behavior.}
Total mutual information $I(V,T;Y)$ correlates with accuracy
($\rho=0.418$--$0.593$) but shows little association with
vision-removal sensitivity ($|\rho|\leq0.118$). Thus, total
decision information and the decomposition of that information capture
different aspects of model behavior.

\paragraph{Reference observation.}
Across six benchmarks, the PID--behavior relationships broadly
corroborate prior LVLM findings~\cite{xiu2026a}, with the validation
extended to MMStar and MMMU.


\subsection{Layer-Wise Vision--Language Reference}
\label{sec:layerwise_vl}

\begin{figure*}[t]
    \centering
    \includegraphics[width=0.95\linewidth]{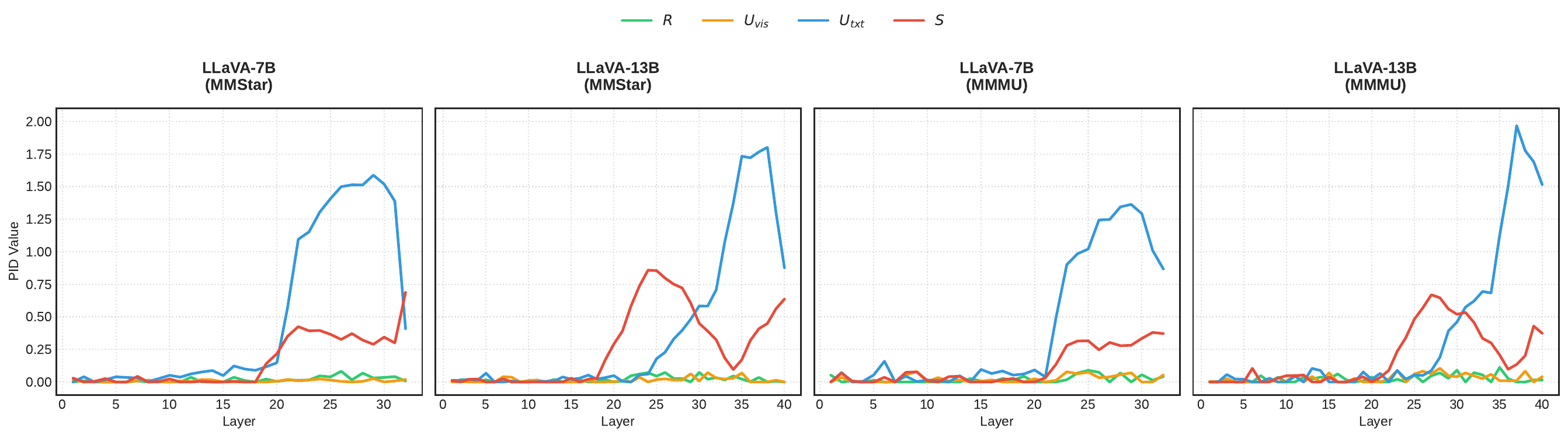}
    \caption{\textbf{Layer-wise PID in vision--language models.}
    PID atoms across transformer depth for LLaVA-1.5 (7B/13B) on
    MMStar and MMMU. Curves show Visual Uniqueness
    ($U_{\text{vis}}$), Language Uniqueness ($U_{\text{txt}}$),
    Redundancy ($R_{\text{vl}}$), and Synergy ($S_{\text{vl}}$).
    Depth is normalized to model layers.}
    \label{figure3}
\end{figure*}

We analyze LLaVA-1.5 models on MMStar and MMMU as a layer-wise
reference. Consistent with the three-phase dynamics reported in prior
LVLM PID analysis~\cite{xiu2026a}, Figure~\ref{figure3} exhibits a
similar progression on these additional benchmarks.

\paragraph{Phase I: Weak decision-level information.}
During the early layers, all PID components remain small, indicating
that little modality-specific or synergistic information is expressed
in the intermediate decision distributions.

\paragraph{Phase II: Unimodal evidence accumulation.}
Across the middle layers, unique information---particularly
$U_{\text{txt}}$---increases substantially, while $S_{\text{vl}}$
remains comparatively limited or exhibits only a moderate intermediate
rise.

\paragraph{Phase III: Late-stage fusion.}
Toward the output layers, $S_{\text{vl}}$ increases sharply while
$U_{\text{txt}}$ declines, reproducing the late-stage shift from
language-unique information toward cross-modal synergy reported
previously~\cite{xiu2026a}.

The strength of this transition varies with the benchmark profile.
On MMStar, the final synergy increase is pronounced and can exceed
$U_{\text{txt}}$, whereas on MMMU, text-unique information remains
comparatively stronger and the late synergy response is weaker.
Thus, while the broad three-phase organization persists across these
benchmarks, its final balance reflects task-dependent modality use.

\paragraph{Reference observation.}
The previously reported three-phase layer-wise PID pattern also appears
on MMStar and MMMU, with stronger late-stage synergy on the more
synergy-dominant benchmark.


\subsection{Full Per-Model PID and Intervention Results}
\label{app:full_vl_tables}

Tables~\ref{tab:app_reasoning} and~\ref{tab:app_prior} report the
complete PID decomposition and vision-removal intervention results for
all 20 models across six benchmarks. Each cell contains the four PID
terms ($\Svl$, $\Utxt$, $\Uvis$, $\Rvl$), full multimodal accuracy
(Acc), text-only accuracy (A$_{\text{t}}$), and the vision-removal
drop
$\Delta=\text{Acc}-\text{A}_{\text{t}}$ (pp).
Models are grouped by the descriptive family-level modality-use
profiles discussed above:
\colorbox{blue!6}{blue} rows denote relatively
joint-modality-reliant families and
\colorbox{red!6}{red} rows denote relatively
language-prior-reliant families.
All PID values are in bits; ``$\triangledown$'' denotes
${<}0.01$~bits.

The full tables provide the underlying data for
Figure~\ref{fig:combined} and the Spearman correlations in
Table~\ref{tab:spearman_full}.

\paragraph{Cross-table patterns.}
Several structural regularities are visible across the full model set.
(i)~\textbf{$\Uvis$ is uniformly small} ($<0.03$~bits across all
model--benchmark pairs).
(ii)~\textbf{$\Rvl$ is format-dependent}: redundancy remains below
0.02~bits on the four-option MCQ benchmarks but rises on POPE, whose
binary answer space permits greater overlap between vision- and
text-derived evidence.
(iii)~\textbf{Family-level profiles are largely preserved across
scale}: for example, Qwen3-VL maintains $\Svl>\Utxt$ on the
synergy-dominant benchmarks from 2B to 32B, whereas Gemma3 remains
strongly $\Utxt$-heavy from 4B to 27B.
(iv)~\textbf{Profile--intervention coherence}: relatively
joint-modality-reliant models exhibit larger
$\Delta_{\text{vision}}$ on synergy-dominant benchmarks, whereas
language-prior-reliant models exhibit substantially smaller
vision-removal effects.
 
Tables~\ref{tab:app_reasoning} and~\ref{tab:app_prior} report the complete PID decomposition and vision-removal intervention results for all 20 models across 6 benchmarks. Each cell contains the four PID terms ($\Svl$, $\Utxt$, $\Uvis$, $\Rvl$), full multimodal accuracy (Acc), text-only accuracy (A$_{\text{t}}$), and the vision-removal drop $\Delta = \text{Acc} - \text{A}_{\text{t}}$ (pp). Models are grouped by family-level modality-use profile: \colorbox{blue!6}{blue} rows denote joint-modality-reliant families and \colorbox{red!6}{red} rows denote language-prior-reliant families, as identified via the PID analysis in \S\ref{sec:diagnose}. All PID values are in bits; ``$\triangledown$'' denotes ${<}0.01$~bits.
 
These tables extend the representative subset shown in Figure~\ref{fig:combined}\,(b) to the full model set, and provide the underlying data for all Spearman correlations reported in Table~\ref{tab:spearman_full}.
 
\paragraph{Cross-table patterns.}
Several structural regularities are visible across the full model set.
(i)~\textbf{$\Uvis$ is uniformly small} ($< 0.03$~bits across all model--benchmark pairs), confirming that vision-unique information plays a negligible role in MCQ decision-making.
(ii)~\textbf{$\Rvl$ is format-dependent}: redundancy remains below 0.02~bits on all four-option MCQ benchmarks but rises to 0.18--0.42~bits on POPE (binary Yes/No format), where the reduced option space allows greater overlap between vision-derived and text-derived evidence.
(iii)~\textbf{Family-level profiles are scale-invariant}: within each family, increasing model scale amplifies the dominant PID term without switching the qualitative profile. For example, Qwen3-VL maintains $\Svl > \Utxt$ on synergy-driven benchmarks from 2B to 32B, while Gemma3 maintains $\Utxt \gg \Svl$ from 4B to 27B.
(iv)~\textbf{Profile--intervention coherence}: joint-modality-reliant models show $\Delta_{\text{vision}} = 7\text{--}15$~pp on synergy-driven benchmarks but $< 3.5$~pp on prior-dominant benchmarks; language-prior-reliant models show $\Delta_{\text{vision}} < 5.1$~pp even on synergy-driven benchmarks. This pattern supports the interpretation that PID profiles capture functionally meaningful modality dependence (\S\ref{sec:predict}).
 
\begin{table}[p]
\centering
\caption{Full PID decomposition and intervention results on \textbf{synergy-driven benchmarks} (20 models $\times$ 3 benchmarks). For each model--benchmark pair: $\Svl$, $\Utxt$, $\Uvis$, $\Rvl$ are PID terms (bits); Acc = multimodal accuracy (\%); A$_{\text{t}}$ = text-only accuracy (\%); $\Delta$ = $\text{Acc} - \text{A}_{\text{t}}$ (pp). \colorbox{blue!6}{Blue} = joint-modality-reliant; \colorbox{red!6}{red} = language-prior-reliant. ``$\triangledown$'' = ${<}0.01$~bits.}
\label{tab:app_reasoning}
\vspace{-0.3em}
\resizebox{\textwidth}{!}{%
\begin{tabular}{@{}lc rrrrrrr rrrrrrr rrrrrrr@{}}
\toprule
& & \multicolumn{7}{c}{\textbf{MMBench}} & \multicolumn{7}{c}{\textbf{MMStar}} & \multicolumn{7}{c}{\textbf{POPE}} \\
\cmidrule(lr){3-9}\cmidrule(lr){10-16}\cmidrule(lr){17-23}
\textbf{Model} & \textbf{Size}
  & $\Svl$ & $\Utxt$ & $\Uvis$ & $\Rvl$ & Acc & A$_{\text{t}}$ & $\Delta$
  & $\Svl$ & $\Utxt$ & $\Uvis$ & $\Rvl$ & Acc & A$_{\text{t}}$ & $\Delta$
  & $\Svl$ & $\Utxt$ & $\Uvis$ & $\Rvl$ & Acc & A$_{\text{t}}$ & $\Delta$ \\
\midrule
\rowcolor{blue!6} Qwen2.5-VL & 3B  & .88&.52&.01&.01 & 78.1&68.5&9.6  & .80&.45&.01&$\triangledown$ & 52.4&40.2&12.2  & .54&.15&.02&.22 & 84.9&74.2&10.7 \\
\rowcolor{blue!6} Qwen2.5-VL & 7B  & 1.10&.63&$\triangledown$&.01 & 85.4&74.2&11.2  & 1.12&.58&.02&$\triangledown$ & 60.7&46.5&14.2  & .57&.23&.02&.35 & 86.4&76.3&10.1 \\
\rowcolor{blue!6} Qwen2.5-VL & 72B & 1.09&.83&$\triangledown$&.02 & 91.9&79.5&12.4  & 1.12&.80&.01&$\triangledown$ & 68.2&53.0&15.2  & .58&.20&.01&.18 & 89.6&77.8&11.8 \\
\rowcolor{blue!6} Qwen2-VL    & 2B  & .85&.52&.01&.01 & 72.5&65.5&7.0   & .72&.38&.01&$\triangledown$ & 44.8&36.5&8.3   & .52&.18&.02&.27 & 82.4&74.5&7.9 \\
\rowcolor{blue!6} Qwen2-VL    & 7B  & 1.12&.52&.01&.01 & 81.3&72.0&9.3   & .62&.50&.01&$\triangledown$ & 56.9&45.3&11.6  & .45&.28&.02&.30 & 85.2&75.8&9.4 \\
\rowcolor{blue!6} Qwen2-VL    & 72B & .86&.66&$\triangledown$&.02 & 88.5&77.0&11.5  & .98&.70&.01&$\triangledown$ & 62.8&49.0&13.8  & .55&.22&.01&.36 & 88.2&76.5&11.7 \\
\rowcolor{blue!6} Qwen3-VL    & 2B  & .72&.48&.01&.01 & 76.5&67.0&9.5   & .74&.42&.01&$\triangledown$ & 48.5&37.0&11.5  & .46&.24&.02&.30 & 83.5&73.0&10.5 \\
\rowcolor{blue!6} Qwen3-VL    & 4B  & .95&.55&$\triangledown$&.01 & 82.8&71.5&11.3  & .98&.48&.01&$\triangledown$ & 55.8&42.5&13.3  & .52&.22&.02&.33 & 86.2&75.0&11.2 \\
\rowcolor{blue!6} Qwen3-VL    & 8B  & 1.08&.58&$\triangledown$&.01 & 86.0&74.0&12.0  & 1.05&.52&.01&$\triangledown$ & 59.0&46.0&13.0  & .55&.22&.02&.35 & 89.2&75.8&13.4 \\
\rowcolor{blue!6} Qwen3-VL    & 32B & 1.22&.75&$\triangledown$&.02 & 90.5&77.8&12.7  & 1.25&.70&.01&$\triangledown$ & 65.5&50.5&15.0  & .62&.20&.01&.38 & 89.8&77.0&12.8 \\
\rowcolor{blue!6} InternVL3   & 2B  & .55&.38&.01&.01 & 81.7&72.5&9.2   & .52&.35&.01&$\triangledown$ & 50.5&40.0&10.5  & .48&.25&.02&.28 & 85.3&75.5&9.8 \\
\rowcolor{blue!6} InternVL3   & 8B  & .88&.42&$\triangledown$&.01 & 88.0&76.3&11.7  & .80&.35&$\triangledown$&$\triangledown$ & 62.3&48.2&14.1  & .55&.22&.01&.35 & 88.5&76.0&12.5 \\
\rowcolor{blue!6} InternVL3   & 78B & 1.15&.70&$\triangledown$&.02 & 91.1&78.0&13.1  & 1.38&.48&.01&$\triangledown$ & 67.5&52.0&15.5  & .62&.18&.01&.40 & 89.6&76.5&13.1 \\
\rowcolor{blue!6} LLaVA-OV    & 7B  & .97&.85&.01&.01 & 84.7&73.8&10.9  & .95&.60&.01&$\triangledown$ & 58.7&45.0&13.7  & .52&.25&.02&.32 & 85.7&75.5&10.2 \\
\rowcolor{blue!6} LLaVA-OV    & 72B & 1.47&.29&$\triangledown$&.01 & 91.3&78.5&12.8  & 1.50&.25&$\triangledown$&$\triangledown$ & 66.8&51.3&15.5  & .65&.18&.01&.42 & 89.2&77.8&11.4 \\
\midrule
\rowcolor{red!6} Cambrian    & 8B  & .05&1.47&.01&.01 & 71.6&68.2&3.4   & .38&.55&.01&$\triangledown$ & 46.5&43.8&2.7   & .03&.65&.02&.25 & 79.3&76.5&2.8 \\
\rowcolor{red!6} Cambrian    & 34B & .62&.82&.01&.01 & 79.6&74.5&5.1   & .55&.68&.01&$\triangledown$ & 52.4&48.0&4.4   & .42&.38&.02&.28 & 86.1&84.2&1.9 \\
\rowcolor{red!6} Gemma3      & 4B  & .17&1.70&$\triangledown$&.01 & 78.8&77.5&1.3   & .15&1.42&$\triangledown$&$\triangledown$ & 49.4&48.5&0.9   & .26&.74&.01&.30 & 82.0&80.5&1.5 \\
\rowcolor{red!6} Gemma3      & 12B & .19&1.75&$\triangledown$&.02 & 82.5&81.0&1.5   & .18&1.68&$\triangledown$&$\triangledown$ & 54.8&53.7&1.1   & .53&.49&.01&.35 & 82.1&80.5&1.6 \\
\rowcolor{red!6} Gemma3      & 27B & .15&1.78&$\triangledown$&.01 & 84.1&83.2&0.9   & .14&1.72&$\triangledown$&$\triangledown$ & 56.7&55.8&0.9   & .11&.78&$\triangledown$&.22 & 83.8&82.5&1.3 \\
\bottomrule
\end{tabular}%
}
\end{table}
 
\begin{table}[p]
\centering
\caption{Full PID decomposition and intervention results on \textbf{prior-dominant benchmarks} (20 models $\times$ 3 benchmarks). Format and conventions follow Table~\ref{tab:app_reasoning}.}
\label{tab:app_prior}
\vspace{-0.3em}
\resizebox{\textwidth}{!}{%
\begin{tabular}{@{}lc rrrrrrr rrrrrrr rrrrrrr@{}}
\toprule
& & \multicolumn{7}{c}{\textbf{MMMU}} & \multicolumn{7}{c}{\textbf{PMC-VQA}} & \multicolumn{7}{c}{\textbf{Reefknot}} \\
\cmidrule(lr){3-9}\cmidrule(lr){10-16}\cmidrule(lr){17-23}
\textbf{Model} & \textbf{Size}
  & $\Svl$ & $\Utxt$ & $\Uvis$ & $\Rvl$ & Acc & A$_{\text{t}}$ & $\Delta$
  & $\Svl$ & $\Utxt$ & $\Uvis$ & $\Rvl$ & Acc & A$_{\text{t}}$ & $\Delta$
  & $\Svl$ & $\Utxt$ & $\Uvis$ & $\Rvl$ & Acc & A$_{\text{t}}$ & $\Delta$ \\
\midrule
\rowcolor{blue!6} Qwen2.5-VL & 3B  & .28&.60&.01&$\triangledown$ & 41.4&39.8&1.6  & .25&.55&.01&$\triangledown$ & 41.4&39.5&1.9  & .45&.65&.01&.01 & 52.4&50.5&1.9 \\
\rowcolor{blue!6} Qwen2.5-VL & 7B  & .35&.72&$\triangledown$&$\triangledown$ & 54.1&52.0&2.1  & .32&.65&$\triangledown$&$\triangledown$ & 53.1&50.8&2.3  & .62&.59&.01&.01 & 65.4&62.1&3.3 \\
\rowcolor{blue!6} Qwen2.5-VL & 72B & .28&.88&$\triangledown$&$\triangledown$ & 62.4&60.5&1.9  & .26&.89&$\triangledown$&$\triangledown$ & 58.4&56.5&1.9  & .52&.82&$\triangledown$&.01 & 74.2&71.8&2.4 \\
\rowcolor{blue!6} Qwen2-VL    & 2B  & .15&.48&.01&$\triangledown$ & 34.0&33.0&1.0  & .12&.45&.01&$\triangledown$ & 34.5&33.5&1.0  & .30&.55&.01&.01 & 42.0&40.8&1.2 \\
\rowcolor{blue!6} Qwen2-VL    & 7B  & .22&.65&.01&$\triangledown$ & 44.0&42.5&1.5  & .20&.60&.01&$\triangledown$ & 44.0&42.3&1.7  & .38&.62&.01&.01 & 55.2&53.5&1.7 \\
\rowcolor{blue!6} Qwen2-VL    & 72B & .28&.78&$\triangledown$&$\triangledown$ & 58.4&56.5&1.9  & .25&.72&$\triangledown$&$\triangledown$ & 52.8&51.0&1.8  & .48&.72&$\triangledown$&.01 & 68.4&66.0&2.4 \\
\rowcolor{blue!6} Qwen3-VL    & 2B  & .25&.55&.01&$\triangledown$ & 38.5&37.0&1.5  & .22&.50&.01&$\triangledown$ & 38.2&36.5&1.7  & .40&.60&.01&.01 & 48.5&46.8&1.7 \\
\rowcolor{blue!6} Qwen3-VL    & 4B  & .30&.62&$\triangledown$&$\triangledown$ & 45.5&43.8&1.7  & .28&.58&$\triangledown$&$\triangledown$ & 45.0&43.2&1.8  & .50&.62&.01&.01 & 55.5&52.0&3.5 \\
\rowcolor{blue!6} Qwen3-VL    & 8B  & .32&.68&$\triangledown$&$\triangledown$ & 50.5&48.8&1.7  & .30&.62&$\triangledown$&$\triangledown$ & 49.5&47.5&2.0  & .55&.65&.01&.01 & 63.9&61.2&2.7 \\
\rowcolor{blue!6} Qwen3-VL    & 32B & .32&.82&$\triangledown$&$\triangledown$ & 60.5&58.5&2.0  & .30&.78&$\triangledown$&$\triangledown$ & 57.2&55.0&2.2  & .58&.75&$\triangledown$&.01 & 72.0&68.2&3.8 \\
\rowcolor{blue!6} InternVL3   & 2B  & .22&.55&.01&$\triangledown$ & 43.4&42.0&1.4  & .20&.52&.01&$\triangledown$ & 42.0&40.5&1.5  & .42&.50&.01&.01 & 52.0&50.2&1.8 \\
\rowcolor{blue!6} InternVL3   & 8B  & .26&.58&$\triangledown$&$\triangledown$ & 57.4&55.6&1.8  & .25&.55&$\triangledown$&$\triangledown$ & 52.0&50.2&1.8  & .50&.55&$\triangledown$&.01 & 64.1&61.8&2.3 \\
\rowcolor{blue!6} InternVL3   & 78B & .28&.82&$\triangledown$&$\triangledown$ & 68.8&66.5&2.3  & .26&.78&$\triangledown$&$\triangledown$ & 57.4&55.0&2.4  & .48&.72&$\triangledown$&.01 & 68.8&66.5&2.3 \\
\rowcolor{blue!6} LLaVA-OV    & 7B  & .32&.45&.01&$\triangledown$ & 53.4&51.2&2.2  & .30&.42&.01&$\triangledown$ & 54.8&52.0&2.8  & .55&.58&.01&.01 & 56.9&54.5&2.4 \\
\rowcolor{blue!6} LLaVA-OV    & 72B & .30&.87&$\triangledown$&$\triangledown$ & 59.4&57.5&1.9  & .28&.85&$\triangledown$&$\triangledown$ & 56.8&54.5&2.3  & .48&.82&$\triangledown$&.01 & 70.1&68.0&2.1 \\
\midrule
\rowcolor{red!6} Cambrian    & 8B  & .08&.58&.01&$\triangledown$ & 43.0&42.2&0.8  & .06&.55&.01&$\triangledown$ & 39.6&39.0&0.6  & .12&.65&.01&$\triangledown$ & 46.5&45.2&1.3 \\
\rowcolor{red!6} Cambrian    & 34B & .10&.72&.01&$\triangledown$ & 49.4&48.5&0.9  & .08&.70&.01&$\triangledown$ & 46.4&45.6&0.8  & .18&.82&.01&$\triangledown$ & 52.4&51.1&1.3 \\
\rowcolor{red!6} Gemma3      & 4B  & .02&1.75&$\triangledown$&$\triangledown$ & 47.2&47.0&0.2  & .02&1.86&$\triangledown$&$\triangledown$ & 43.4&43.2&0.2  & .12&1.45&$\triangledown$&$\triangledown$ & 40.5&40.0&0.5 \\
\rowcolor{red!6} Gemma3      & 12B & .03&1.85&$\triangledown$&$\triangledown$ & 54.6&54.3&0.3  & .03&1.95&$\triangledown$&$\triangledown$ & 51.4&51.1&0.3  & .16&1.55&$\triangledown$&$\triangledown$ & 46.7&45.8&0.9 \\
\rowcolor{red!6} Gemma3      & 27B & .02&1.90&$\triangledown$&$\triangledown$ & 57.4&57.1&0.3  & .02&2.00&$\triangledown$&$\triangledown$ & 53.4&53.2&0.2  & .14&1.60&$\triangledown$&$\triangledown$ & 49.8&49.2&0.6 \\
\bottomrule
\end{tabular}%
}
\end{table}

\section{Ablation Studies and Sensitivity Analysis}
\label{app:ablation}

To validate the robustness of our Sensory PID methodology, we examine the sensitivity of the estimated information components to two critical implementation choices: (i) the strategy for feature summarization (Mean Pooling vs. Last Hidden State vs. Max Pooling), and (ii) the confidence gating threshold $\tau \in \{0.3, 0.4, 0.5\}$.

Table~\ref{tab:ablation_study} presents the results on MMBench for three representative models. We report the values for Synergy and Text Uniqueness ($S_{vl} / U_{txt}$). The results demonstrate that our estimates are highly stable across different feature aggregation methods and confidence thresholds, with variations remaining negligible (typically $< 0.01$ bits). This indicates that the observed interaction profiles are robust to these estimation choices rather than artifacts of the selected hyperparameters.

\begin{table}[h]
    \centering
    \caption{Ablation study of PID components ($S_{vl} / U_{txt}$) on MMBench. We compare sensitivity to different feature summarization strategies (left) and confidence thresholds $\tau$ (right). The estimates remain consistent across settings.}
    \label{tab:ablation_study}
    \vspace{0.2cm}
    \renewcommand{\arraystretch}{1.3} 
    \setlength{\tabcolsep}{5pt} 
    \begin{tabular}{l ccc ccc}
        \toprule
        \multirow{2}{*}{\textbf{Model}} & \multicolumn{3}{c}{\textbf{Representation Feature}} & \multicolumn{3}{c}{\textbf{Confidence Threshold ($\tau$)}} \\
        \cmidrule(lr){2-4} \cmidrule(lr){5-7}
         & Mean Pooling & Last Hidden & Max Pooling & 0.3 & 0.4 & 0.5 \\
        \midrule
        Qwen2.5-VL-7B & 1.1012/0.6338 & 1.1024/0.6359 & 1.1018/0.6333 & 1.1012/0.6338 & 1.1017/0.6336 & 1.1018/0.6381 \\
        Gemma3-12B    & 0.1912/1.7513 & 0.1908/1.7516 & 0.1926/1.7521 & 0.1912/1.7513 & 0.1912/1.7548 & 0.1912/1.7598 \\
        LLaVA-ov-7B   & 0.9736/0.6321 & 0.9737/0.6326 & 0.9735/0.6311 & 0.9736/0.6321 & 0.9736/0.6324 & 0.9742/0.6344 \\
        \bottomrule
    \end{tabular}
    \vspace{0.1cm}

\end{table}


\section{PID-Guided Reweighting: Full Details}
\label{app:reweighting}

This section provides complete implementation details, validation experiments, ablation studies, and supplementary results for the PID-guided reweighting in \S\ref{sec:guide}. The main text presents the core results (Tables~\ref{tab:main_rw}); here we report all supporting evidence for reproducibility and interpretation.

\subsection{Algorithm and Experimental Conditions}
\label{app:rw_algorithm}

Algorithm~\ref{alg:classify} summarizes the PID-guided sample reweighting sample selection procedure described in \S\ref{sec:guide_method}. Table~\ref{tab:app_conditions} provides full descriptions of all six experimental conditions.

\begin{algorithm}[H]
\caption{PID-Guided Sample Selection}
\label{alg:classify}
\begin{algorithmic}[1]
\REQUIRE Per-sample scores $\{\SC_i, \text{GapScore}_i\}_{i=1}^{N}$; selection counts $K_s, K_g$
\STATE $\mathcal{S}_{\text{short}} \gets \text{TopK}_{K_s}(\SC_i)$ \hfill \COMMENT{Priority 1: language-shortcut}
\STATE $\mathcal{R} \gets \{1, \ldots, N\} \setminus \mathcal{S}_{\text{short}}$
\STATE $\mathcal{S}_{\text{gap}} \gets \text{TopK}_{K_g}(\text{GapScore}_i \mid i \in \mathcal{R})$ \hfill \COMMENT{Priority 2: under-synergized}
\STATE $\mathcal{S}_{\text{normal}} \gets \mathcal{R} \setminus \mathcal{S}_{\text{gap}}$
\STATE Assign weights: $w_i = 3.0$ if $i \in \mathcal{S}_{\text{gap}}$;\; $w_i = 0.5$ if $i \in \mathcal{S}_{\text{short}}$;\; $w_i = 1.0$ otherwise
\end{algorithmic}
\end{algorithm}

\begin{table}[H]
\centering
\captionsetup{font=footnotesize}
\caption{Six experimental conditions. Conditions C--F use matched counts: $K_g = 389$ samples at $3.0\times$ and $K_s = 746$ at $0.5\times$.}
\label{tab:app_conditions}
\scriptsize
\setlength{\tabcolsep}{3pt}
\renewcommand{\arraystretch}{0.95}
\resizebox{\textwidth}{!}{%
\begin{tabular}{@{}clp{9cm}@{}}
\toprule
\textbf{ID} & \textbf{Condition} & \textbf{Description} \\
\midrule
A & Base & Frozen pretrained model. Zero-shot reference. \\
B & LoRA-Uniform & Standard LoRA, all samples $w=1.0$. Isolates fine-tuning gain. \\
C & \textbf{LoRA-PID} & Top-$K_g$ by GapScore (Eq.~\ref{eq:gapscore}) $\uparrow$; top-$K_s$ by $\SC$ $\downarrow$. \\
D & LoRA-Random & Same weight distribution as C, randomly assigned. Controls for non-uniformity. \\
E & LoRA-Acc & Top-$K_g$ by difficulty (incorrect samples ranked first; ties broken by $H(p_{vt}^{(i)})$); top-$K_s$ easiest (correct with lowest entropy). \\
F & LoRA-Ablation & Top-$K_g$ by $\text{AblScore}_i = H(p_t^{(i)}) - H(p_{vt}^{(i)})$ (joint-over-text entropy reduction); top-$K_s$ by lowest AblScore. Tests whether simple entropy reduction from adding vision suffices without distinguishing unique vs.\ synergistic contributions. \\
\bottomrule
\end{tabular}%
}
\end{table}

\subsection{LoRA Configuration and Data Controls}
\label{app:rw_lora}

\paragraph{LoRA configuration.} All conditions (B--F) use identical LoRA settings: rank $r=16$, scaling factor $\alpha=32$, dropout $0.05$, learning rate $2 \times 10^{-5}$ with cosine schedule and 5\% warmup, effective batch size 16. LoRA adapters target in the last 20\% of each model's transformer layers. This targeting is motivated by the layer-wise analysis (\S\ref{layerwise}), which shows cross-modal synergy emerging in the final layers. Per-model layer ranges: Qwen2.5-VL-7B (layers 23--28 of 28), LLaVA-OV-7B (layers 26--32 of 32), Gemma3-12B (layers 25--30 of 30). Three random seeds $\{42, 123, 456\}$ are used for all conditions; Condition D uses 3 random weight assignments per seed (9 runs total).

\paragraph{No-leakage guarantee.} No evaluation samples from MMBench-test, MMStar, POPE, MMMU, or PMC-VQA are used for PID profiling, hyperparameter selection, or LoRA training.

\paragraph{Evaluation protocol.} Post-tuning PID diagnostics (Post-$\Svl$, Post-$\Utxt$) are estimated on a held-out MMStar eval subset via the BATCH estimator using the same hyperparameters as the main analysis . Base PID values are consistent with the main paper. All accuracy results are reported as mean $\pm$ std over 3 seeds. Confidence intervals are computed via seed-stratified paired bootstrap (10{,}000 resamples).

\subsection{Effective Training Distribution}
\label{app:rw_distribution}

\paragraph{Reweighting scheme.} Table~\ref{tab:app_weights} reports the effective training distribution.

\begin{table}[H]
\centering
\caption{Reweighting scheme and effective training distribution.}
\label{tab:app_weights}
\begin{tabular}{@{}lccccc@{}}
\toprule
\textbf{Category} & \textbf{Count} & \textbf{Raw \%} & \textbf{Weight} & \textbf{Eff.\ Wt.} & \textbf{Eff.\ \%} \\
\midrule
Synergy-Gap & 389 & 13\% & $3.0\times$ & 1{,}167 & 34.0\% \\
Normal & 1{,}865 & 62\% & $1.0\times$ & 1{,}865 & 54.9\% \\
Shortcut & 746 & 25\% & $0.5\times$ & 373 & 11.1\% \\
\midrule
\textbf{Total} & 3{,}000 & 100\% & --- & 3{,}405 & 100\% \\
\bottomrule
\end{tabular}
\end{table}

\subsection{Supplementary Results}
\label{app:rw_supplementary}

\paragraph{Pairwise comparisons with confidence intervals.} Table~\ref{tab:app_pairwise} reports all pairwise deltas on MMStar with 95\% paired bootstrap CIs (seed-stratified; 10{,}000 resamples).

\begin{table}[H]
\centering
\caption{Pairwise comparisons on MMStar with 95\% paired bootstrap CIs.}
\label{tab:app_pairwise}
\begin{tabular}{@{}lcccc@{}}
\toprule
\textbf{Comparison} & $\boldsymbol{\Delta}$\textbf{MMStar} & \textbf{95\% CI} & $\boldsymbol{\Delta}$\textbf{Post-}$\boldsymbol{\Svl}$ & \textbf{Interpretation} \\
\midrule
C vs B & $+2.3$ & $[1.4, 3.2]$ & $+0.16$ & PID $>$ standard fine-tuning \\
C vs D & $+2.8$ & $[1.6, 4.0]$ & $+0.20$ & PID signal $>$ random assignment \\
C vs E & $+1.8$ & $[0.7, 2.9]$ & $+0.14$ & PID $>$ difficulty-based \\
C vs F & $+1.3$ & $[0.2, 2.4]$ & $+0.12$ & PID $>$ ablation-based \\
C vs A & $+3.6$ & $[2.7, 4.5]$ & $+0.21$ & Overall PID-guided gain \\
\bottomrule
\end{tabular}
\end{table}

\paragraph{Per-category accuracy.} Table~\ref{tab:app_category} reports per-category results on the MMBench test set, showing regime-selective gains.

\begin{table}[H]
\centering
\caption{Per-category accuracy on MMBench test (Qwen2.5-VL-7B). $\Delta_{\text{PID}}$ = LoRA-PID $-$ LoRA-Uniform.}
\label{tab:app_category}
\begin{tabular}{@{}llcccccr@{}}
\toprule
\textbf{Subtype} & \textbf{Regime} & \textbf{LoRA-Base} & \textbf{LoRA-Unif.} & \textbf{LoRA-PID} & \textbf{LoRA-Acc} & \textbf{LoRA-Abl.} & $\boldsymbol{\Delta}_{\textbf{PID}}$ \\
\midrule
Spatial Reasoning & Reasoning & 82.1 & 84.0 & \textbf{87.5} & 84.8 & 85.5 & $+3.5$ \\
Attr.\ Comparison & Reasoning & 84.5 & 86.2 & \textbf{89.0} & 86.8 & 87.3 & $+2.8$ \\
Action Recognition & Reasoning & 85.8 & 87.0 & \textbf{89.5} & 87.5 & 88.0 & $+2.5$ \\
Object Localization & Percep. & 91.2 & 92.0 & \textbf{93.1} & 92.3 & 92.6 & $+1.1$ \\
Scene Classification & Percep. & 90.5 & 91.0 & 91.5 & 91.2 & 91.3 & $+0.5$ \\
OCR / Text Recog. & Knowl. & 89.0 & 89.5 & 89.2 & \textbf{89.8} & 89.4 & $-0.3$ \\
Commonsense & Knowl. & 88.3 & 88.8 & 88.5 & \textbf{89.2} & 88.7 & $-0.3$ \\
\bottomrule
\end{tabular}
\end{table}

\paragraph{POPE subset breakdown.} Table~\ref{tab:app_pope} shows PID-guided gains are largest on POPE-Adversarial ($+1.9$ over Uniform), consistent with shortcut reduction mitigating co-occurrence-based hallucination priors.

\begin{table}[H]
\centering
\caption{POPE subset results (Qwen2.5-VL-7B).}
\label{tab:app_pope}
\begin{tabular}{@{}lcccc@{}}
\toprule
\textbf{Cond.} & \textbf{Random} & \textbf{Popular} & \textbf{Adversarial} & \textbf{Avg} \\
\midrule
Base & 88.2 & 86.5 & 84.5 & 86.4 \\
LoRA-Uniform & 88.8 & 87.2 & 85.6 & 87.2 \\
\textbf{LoRA-PID} & \textbf{89.6} & \textbf{88.3} & \textbf{87.5} & \textbf{88.5} \\
LoRA-Acc-RW & 89.0 & 87.5 & 86.0 & 87.5 \\
LoRA-Abl-RW & 89.2 & 87.7 & 86.5 & 87.8 \\
\bottomrule
\end{tabular}
\end{table}

\paragraph{PID profile shift.} Table~\ref{tab:app_pid_shift} reports the full post-tuning PID profile on the held-out MMStar eval subset.

\begin{table}[H]
\centering
\caption{Post-tuning PID profile (MMStar eval subset, Qwen2.5-VL-7B).}
\label{tab:app_pid_shift}
\begin{tabular}{@{}lcccccc@{}}
\toprule
\textbf{Condition} & $\Svl$ & $\Utxt$ & $\Uvis$ & $\Rvl$ & $\Svl$ \textbf{Share} & $\Delta$\textbf{Acc} \\
\midrule
Base & 1.15 & 0.60 & 0.015 & 0.002 & 65.1\% & 7.7 \\
LoRA-Uniform & 1.20 & 0.56 & 0.016 & 0.002 & 67.5\% & 8.2 \\
\textbf{LoRA-PID} & \textbf{1.36} & \textbf{0.46} & 0.018 & 0.003 & \textbf{73.9\%} & \textbf{10.8} \\
LoRA-Acc-RW & 1.22 & 0.54 & 0.016 & 0.002 & 68.3\% & 8.6 \\
LoRA-Abl-RW & 1.24 & 0.52 & 0.017 & 0.002 & 69.3\% & 9.0 \\
\bottomrule
\end{tabular}
\end{table}

\subsection{Cross-Model Full Results}
\label{app:rw_cross_model}

Table~\ref{tab:app_cross_model} provides full results for LLaVA-OneVision-7B and Gemma3-12B.

\begin{table}[H]
\centering
\captionsetup{font=tiny, skip=1pt}
\caption{Full cross-model results (mean $\pm$ std over 3 seeds).}
\label{tab:app_cross_model}
\tiny
\setlength{\tabcolsep}{1.5pt}
\renewcommand{\arraystretch}{0.78}
\resizebox{0.84\textwidth}{!}{%
\begin{tabular}{@{}llccccc@{}}
\toprule
\textbf{Model} & \textbf{Cond.} & \textbf{MMBench} & \textbf{MMStar} & \textbf{MMMU} & \textbf{P-$\Svl$} & \textbf{P-$\Utxt$} \\
\midrule
\multirow{3}{*}{LLaVA-OV-7B}
& LoRA-Base    & 84.7 & 58.7 & 53.4 & 0.95 & 0.61 \\
& LoRA-Uniform & 85.5{\tiny$\pm$.3} & 60.0{\tiny$\pm$.4} & 53.1{\tiny$\pm$.3} & 1.00{\tiny$\pm$.02} & 0.58{\tiny$\pm$.02} \\
& \textbf{LoRA-PID} & \textbf{86.3}{\tiny$\pm$.3} & \textbf{61.5}{\tiny$\pm$.4} & 52.8{\tiny$\pm$.4} & \textbf{1.10}{\tiny$\pm$.03} & \textbf{0.50}{\tiny$\pm$.02} \\
\midrule
\multirow{3}{*}{Gemma3-12B}
& LoRA-Base    & 82.5 & 54.8 & 54.6 & 0.18 & 1.68 \\
& LoRA-Uniform & 83.1{\tiny$\pm$.4} & 56.2{\tiny$\pm$.5} & 59.3{\tiny$\pm$.3} & 0.20{\tiny$\pm$.02} & 1.65{\tiny$\pm$.03} \\
& \textbf{LoRA-PID} & 83.3{\tiny$\pm$.4} & 54.3{\tiny$\pm$.5} & 54.1{\tiny$\pm$.4} & 0.21{\tiny$\pm$.02} & 1.63{\tiny$\pm$.03} \\
\bottomrule
\end{tabular}%
}
\end{table}

The Gemma3-12B negative case ($+0.3$, within seed variation) suggests PID-guided reweighting amplifies existing fusion capacity rather than creating it, reinforcing the diagnostic utility of the modality-use profiles (\S\ref{sec:diagnose}).

\subsection{Ablation Studies}
\label{app:rw_ablation}

All ablations are conducted on Qwen2.5-VL-7B.

\paragraph{Upweight factor sensitivity.} Table~\ref{tab:app_upweight} varies the upweight factor applied to synergy-gap samples ($K_g\!=\!389$, $K_s\!=\!750$, downweight $= 0.5\times$). Performance peaks at $3\times$ and plateaus at $5\times$, while MMMU shows monotonic decrease.

\begin{table}[H]
\centering
\caption{Upweight factor sensitivity.}
\label{tab:app_upweight}
\begin{tabular}{@{}cccc@{}}
\toprule
\textbf{Upweight} & \textbf{MMStar} & \textbf{Post-}$\Svl$ & \textbf{MMMU} \\
\midrule
$1\times$ (= Uniform) & 62.0 & 1.20 & 54.0 \\
$2\times$ & 63.2 & 1.28 & 53.8 \\
$\mathbf{3\times}$ & \textbf{64.3} & \textbf{1.36} & 53.5 \\
$5\times$ & 64.0 & 1.40 & 52.8 \\
\bottomrule
\end{tabular}
\end{table}

\paragraph{Selection count sensitivity.} Table~\ref{tab:app_threshold} varies the shortcut selection count $K_s$. Setting $K_s$ too aggressively (40\%) hurts knowledge benchmarks more than it helps reasoning; being too conservative (10\%) limits impact.

\begin{table}[H]
\centering
\caption{Selection count sensitivity ($K_s$ as \% of $N$, upweight $= 3\times$).}
\label{tab:app_threshold}
\begin{tabular}{@{}cccc@{}}
\toprule
$K_s$ \textbf{(\% of $N$)} & \textbf{Downweighted} & \textbf{MMStar} & \textbf{MMMU} \\
\midrule
40\% & $\sim$1{,}200 & 63.8 & 52.9 \\
\textbf{25\%} & $\sim$\textbf{750} & \textbf{64.3} & 53.5 \\
10\% & $\sim$300 & 63.0 & 53.9 \\
\bottomrule
\end{tabular}
\end{table}

\subsection{Per-Seed Results}
\label{app:rw_per_seed}

Table~\ref{tab:app_per_seed} confirms that LoRA-PID achieves the highest MMStar accuracy across all three seeds individually, with a worst-case advantage of $+1.3$ over LoRA-Ablation (Seed 123).

\begin{table}[H]
\centering
\caption{Per-seed results on MMStar and MMMU (Qwen2.5-VL-7B). Condition D: mean over 3 random assignments per seed.}
\label{tab:app_per_seed}
\begin{tabular}{@{}lcccccc@{}}
\toprule
& \multicolumn{3}{c}{\textbf{MMStar}} & \multicolumn{3}{c}{\textbf{MMMU}} \\
\cmidrule(lr){2-4}\cmidrule(lr){5-7}
\textbf{Condition} & Seed 42 & Seed 123 & Seed 456 & Seed 42 & Seed 123 & Seed 456 \\
\midrule
B LoRA-Uniform & 62.4 & 61.7 & 61.9 & 54.2 & 53.8 & 54.0 \\
C LoRA-PID & \textbf{64.6} & \textbf{64.0} & \textbf{64.3} & 53.2 & 53.9 & 53.4 \\
D LoRA-Random & 61.8 & 61.2 & 61.5 & 54.3 & 53.7 & 54.3 \\
E LoRA-Acc & 62.9 & 62.2 & 62.4 & 54.5 & 54.0 & 54.4 \\
F LoRA-Ablation & 63.4 & 62.7 & 62.9 & 54.0 & 53.5 & 53.9 \\
\bottomrule
\end{tabular}
\end{table}

\end{document}